\documentclass[journal]{IEEEtran}

\usepackage{cite}

\usepackage[pdftex]{graphicx}

\usepackage[cmex10]{amsmath}

\usepackage[ruled, lined, linesnumbered]{algorithm2e}

\usepackage{array} 

\usepackage{subcaption}
 
\usepackage{fixltx2e}

\usepackage{url}

\usepackage{atbegshi}
\AtBeginDocument{\AtBeginShipoutNext{\AtBeginShipoutDiscard}}

\begin{document}

\setcounter{page}{0}

\title{Increasing Behavioral Complexity for Evolved Virtual Creatures with the ESP Method}
%
%
%


\author{Dan Lessin, Don Fussell, Risto Miikkulainen, Sebastian Risi}%
\thanks{Dan Lessin (corresponding author) and Sebastian Risi are with IT University of Copenhagen, Copenhagen, 2300, Denmark (email: daniel.g.lessin@gmail.com; sebr@itu.dk).}%
\thanks{Don Fussell and Risto Miikkulainen are with the University of Texas at Austin, Austin, Texas, 78712, USA (email: fussell@cs.utexas.edu, risto@cs.utexas.edu).}%

\maketitle

\begin{abstract}
Since their introduction in 1994~\cite{Sims:1994:EVC:192161.192167}, evolved virtual creatures (EVCs) have employed the coevolution of morphology and control to produce high-impact work in multiple fields, including graphics, evolutionary computation, robotics, and artificial life.  However, in contrast to fixed-morphology creatures, there has been no clear increase in the behavioral complexity of EVCs in those two decades.

This paper describes a method for moving beyond this
limit, making use of high-level human input in the form
of a syllabus of intermediate learning tasks---along with
mechanisms for preservation, reuse, and combination of
previously learned tasks.  This method---named \emph{ESP} for
its three components: \emph{encapsulation}, 
\emph{syllabus}, and \emph{pandemonium}---is presented in two complementary versions: \emph{Fast ESP}, which constrains later morphological changes to achieve linear growth in computation time as behavioral complexity is added, and \emph{General ESP}, which allows this restriction to be removed when sufficient computational resources are available.  Experiments demonstrate that the ESP method allows evolved virtual creatures to reach new levels of behavioral complexity in the co-evolution of morphology and control, approximately doubling the previous state of the art. 

\end{abstract}


%

\section{Introduction}

\begin{figure}[htbp!]
  \centering

  \begin{subfigure}{0.45\textwidth}
  \includegraphics[width=0.45\textwidth]{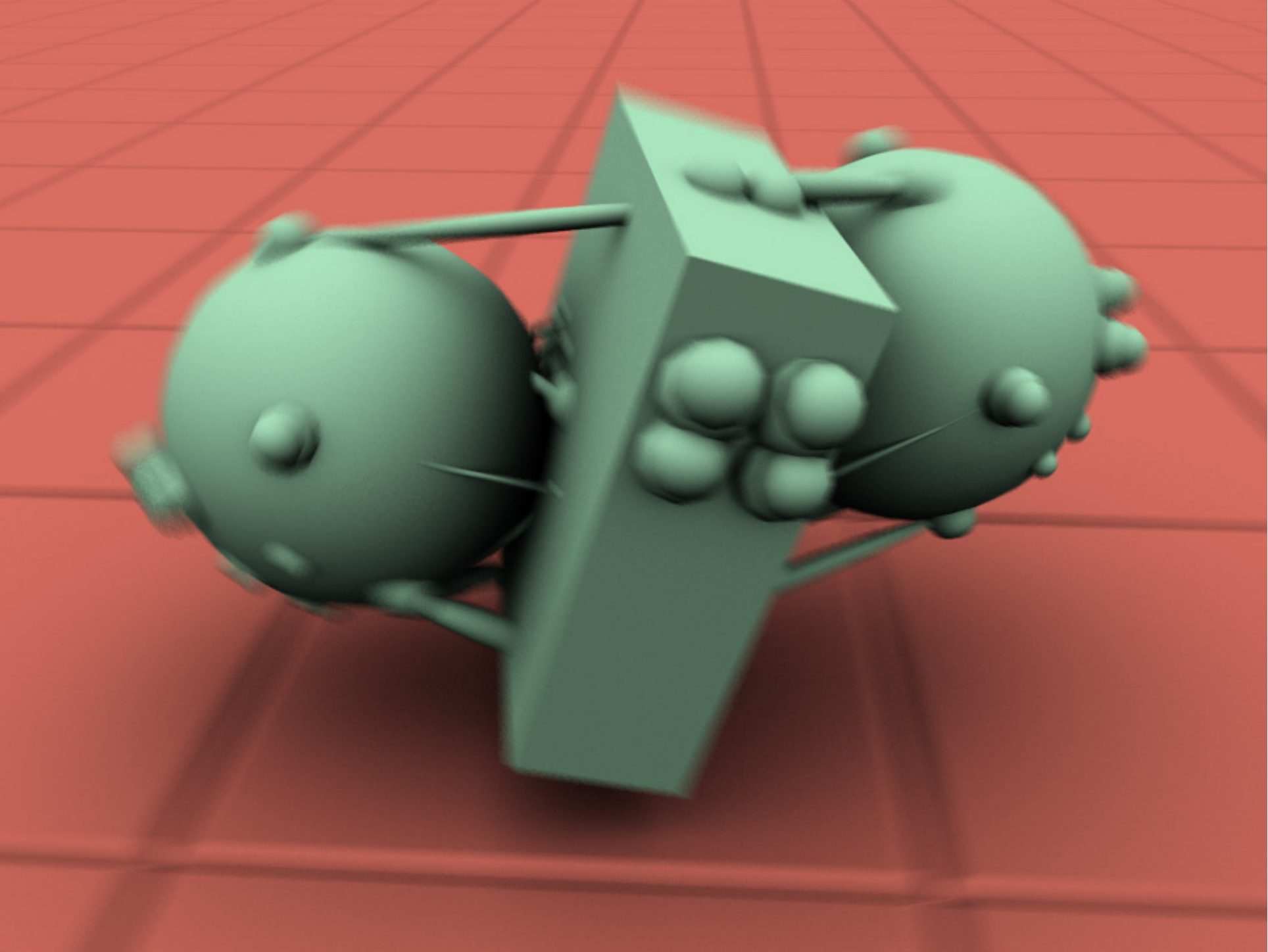}
  \includegraphics[width=0.5\textwidth]{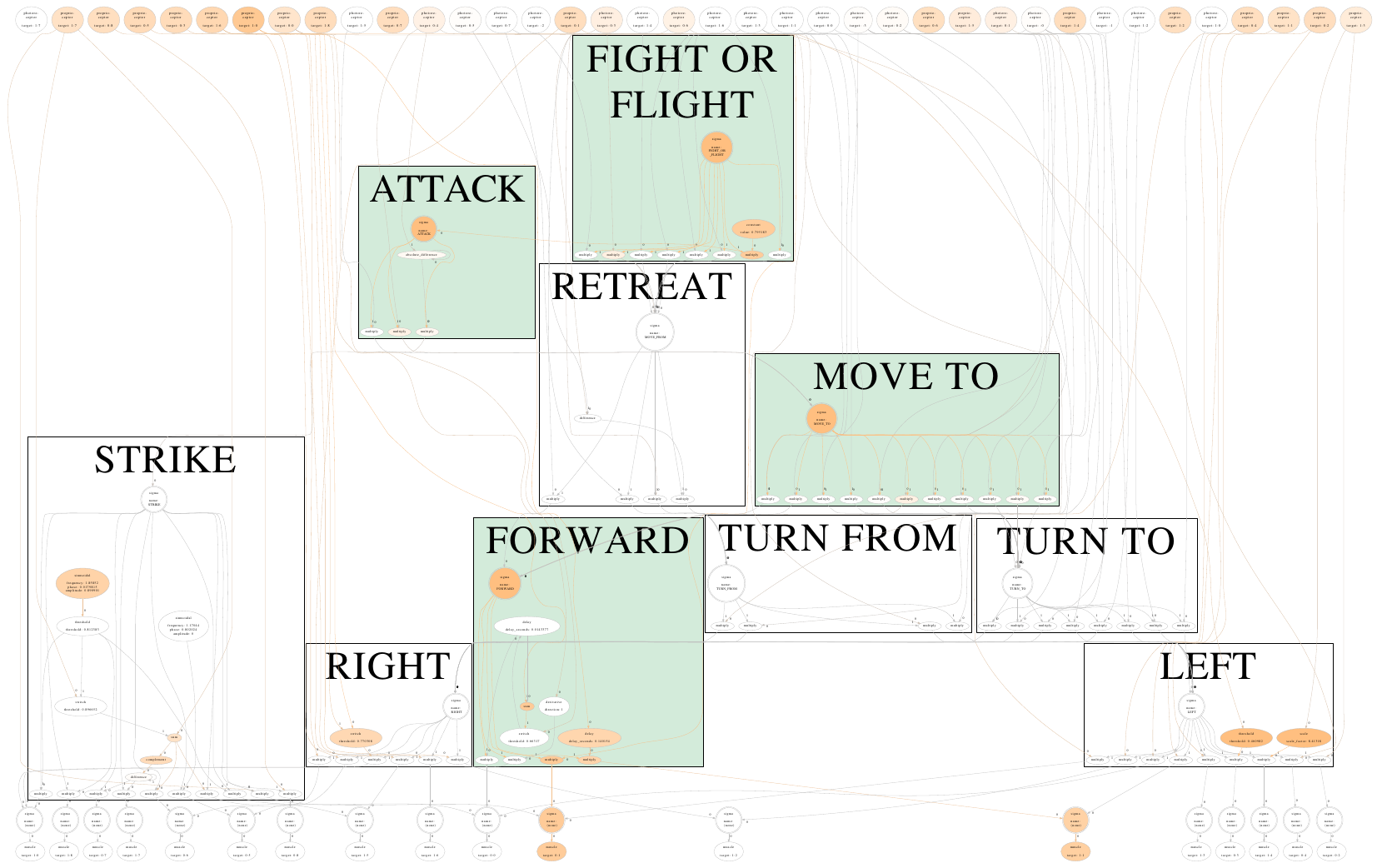}
  \caption{}
  \end{subfigure}
\par\medskip
  \begin{subfigure}{0.45\textwidth}
  \includegraphics[width=\textwidth]{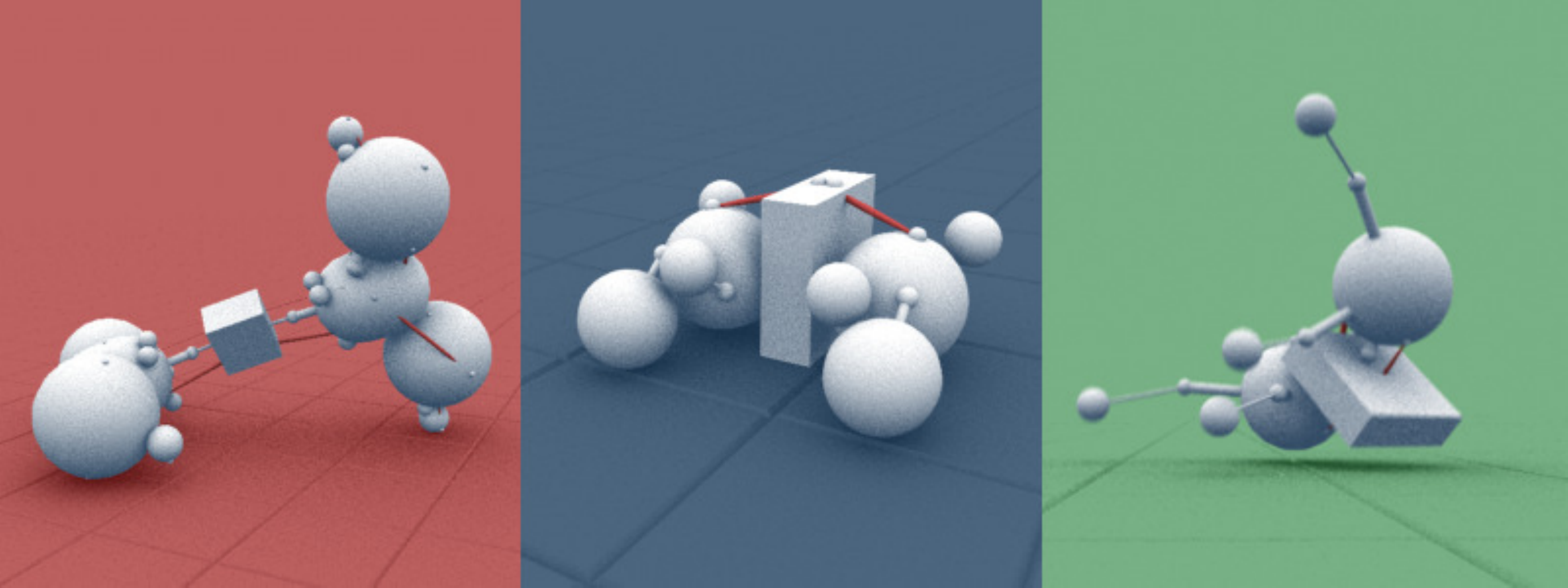}
  \caption{}
  \end{subfigure}

  \caption{
    The two versions of ESP described in this paper.  This paper explores the ESP method for increasing
    behavioral complexity in evolved virtual creatures.  Two implementations of this method are presented: \emph{Fast ESP} and \emph{General ESP}.
    (a) \emph{Fast ESP} allows the open-ended development of behavioral complexity for EVCs with only a linear increase in computational time.  See results produced using Fast ESP at \url{http://youtu.be/dRLNnJlT8rY} .
    (b) When sufficient computational resources are available, \emph{General ESP} permits
    greater body adaptation to multiple tasks, while preserving Fast ESP's ability to produce behavioral complexity.  See results produced using General ESP at \url{http://youtu.be/fyVr7gdGEPE} .
  }
  \label{fig:introduction_overview}
\end{figure}

%
%
%
%

\begin{figure}[ht]
  \setlength{\belowcaptionskip}{-5pt}
  \centering
  \includegraphics[width=0.50\textwidth]{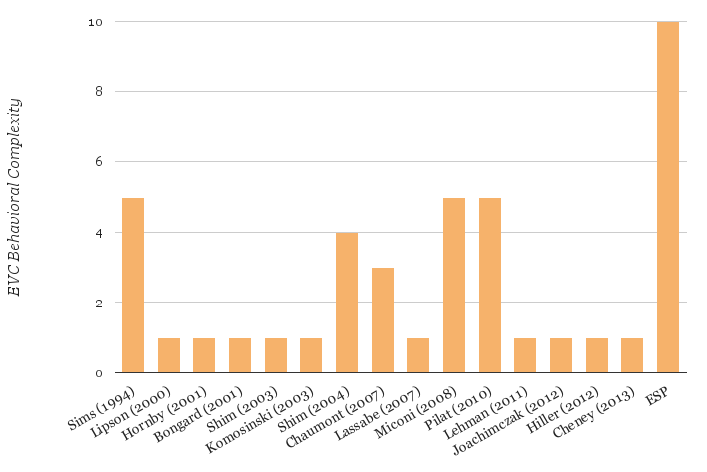}
  \caption{Behavioral complexity in EVCs.  Defined as the number of discriminable behaviors in a creature's repertoire, the behavioral complexity achieved by Sims in 1994 has not been clearly exceeded in later work.  In contrast, the ESP method described in this paper approximately doubles it.}
  \label{fig:behavioral_complexity}
\end{figure}

\IEEEPARstart{D}{efining} \emph{behavioral complexity} as the
number of discriminable
behaviors in a creature's repertoire, most evolved virtual creatures
have minimal complexity,
employing repertoires that contain only a single behavior.
The original examples by Sims~\cite{Sims:1994:EVC:192161.192167} of locomotion on land and in water, as well
as jumping, fall into this category, as does much of the work
that others have since completed.  For example,
locomotion in air~\cite{shim2003generating}, a specialized form
of ground-locomoting EVCs that can be
converted into functional real-world
robots~\cite{Lipson2000},
soft-bodied virtual creatures~\cite{joachimczak2012co,hiller2012automatic}, and many other
variations have this level of complexity~\cite{chaumont2007evolving,hornby2001body,Bongard01repeatedstructure,lassabe2007new,Komosinski-2003,lehman2011evolving}.
The highest level of behavioral complexity 
demonstrated by Sims---creatures with the ability to follow a
target or a path by switching between perhaps up to five
discriminable behaviors---has since been matched multiple
times~\cite{Pilat:2010:EVC:1830483.1830502,shim2004evolving,miconi2008silicon},
but never clearly exceeded (Figure~\ref{fig:behavioral_complexity}).

Yet more complex behaviors would clearly be useful.
Numerous examples of valued
creature content from the real world---nature documentaries,
animal and human combat, even internet cat
videos\footnote{e.g., ``THE BEST CAT VIDEO YOU'LL EVER SEE'' [sic], http://www.youtube.com/watch?v=20mrEtabOLM}---feature more complex behaviors than what has been demonstrated in EVCs to date.
Perhaps if we can bring greater behavioral complexity to EVCs,
they can begin to approach the entertainment value of their
non-virtual counterparts.

In fact, there is suggestive evidence in support of this proposition.
Cognitive science and psychology describe a striking effect in which the
right kinds of relatively complex behaviors---even by the
simplest of geometric figures---lead to the perception of
intentionality and desires (\emph{perceptual
  animacy})~\cite{scholl2000perceptual,heider1944experimental}. For a particularly
clear non-academic example of this same effect, consider the
academy-award-winning animated short ``The Dot and the Line''~\cite{jones_1965}.  In much of this film, the only elements
added to a simple dot and line to transform them into the
protagonists of a compelling love story are their
movements--their behavioral complexity.

Motivated by this potential, this paper describes a method designed to
increase behavioral complexity in evolved
virtual creatures.  The primary elements of this
method, \emph{ESP}---\emph{encapsulation}, \emph{syllabus}, and
\emph{pandemonium}---are defined as follows:

\begin{enumerate}
\item
  A human-designed \emph{syllabus} breaks the development of a
  complex behavior into a sequence of smaller learning tasks.
\item
  Once each of these subskills is learned, it is
  \emph{encapsulated} to preserve it throughout future
  evolution, and also to allow future skills to
  incorporate its function more easily.   
\item
  A mechanism inspired by Selfridge's
  \emph{pandemonium}~\cite{Selfridge1958a} is used to
  resolve disputes between competing skills or drives within
  the increasingly complex brain.
\end{enumerate}

ESP is presented in this paper in two complementary versions: \emph{Fast ESP} and \emph{General ESP}.  By placing some key limitations on morphological changes after the first skill in a syllabus is learned, \emph{Fast ESP} eliminates the need for retesting of previous skills as new skills are added.  This approach allows computational time to grow approximately linearly as behavioral complexity is increased.  In contrast, when sufficient computing resources are available and full morphological adaptation to multiple skills is important, \emph{General ESP} removes the morphological constraints of Fast ESP through a process of retesting and reconciliation.   
This paper offers a unified treatment of both the Fast and General ESP versions, which were first described by Lessin et al. in conference papers~\cite{Lessin:2013:OBC:2463372.2463411} and~\cite{lessin:alife14}, respectively.

In the remainder of this paper, Section~\ref{sec:background} presents the background on EVCs and their typical behaviors.  Section~\ref{sec:underlying_EVC_system} reviews the mechanics of the EVC system underlying the ESP implementation.  In Sections~\ref{sec:ESP} and~\ref{sec:ESP_results} of this paper, Fast ESP is described, and it is employed to
approximately double the state of the art in behavioral
complexity for evolved virtual creatures.  In Sections~\ref{sec:extended_ESP} and~\ref{sec:extended_ESP_results}, General ESP is described, and it is applied to demonstrate a significant increase in
the useful variety and quality of evolved creatures, while still incrementally developing complex behaviors from a
sequence of simpler learning tasks.


\section{Background}
\label{sec:background}

This section provides a review of relevant background material, including EVCs, typical EVC behaviors and their complexity, and task decomposition strategies related to ESP.

\subsection{Evolved Virtual Creatures (EVCs)}

The first and most influential examples of evolved virtual
creatures are due to Sims~\cite{Sims:1994:EVC:192161.192167}.
The genotypes for Sims' creatures were directed graphs, able to
encode complex body structures.  The bodies of
these creatures were composed of boxes, connected by joints
with varying degrees of freedom and evolvable limits to their
revolution.  Actuation was provided by implicit joint motors,
able to apply force at every degree of freedom of every joint.
Sims' brains were composed of nodes computing simple functions, with signals carried
between nodes by evolved connections.  In his
implementation, brain elements may be embedded within body
segments, where they can take advantage of the same kinds of
repetition and recursion as the creature's morphology.
Evolution in Sims' system made
use of fitness-proportionate selection, crossover, mutation,
and elitism.

Using this system, Sims demonstrated impressive results in
multiple tasks.  A variety of creatures were evolved for
locomotion, both on land and in water, and creatures with
the ability to jump off the ground were produced.  Most
impressively, Sims demonstrated creatures evolved for
phototaxis (light seeking) behavior.  Many of these behaviors have since
become benchmarks for EVCs.


\subsection{Locomotion}
\label{sec:background_locomotion}

The standard benchmark task for an EVC system is locomotion.
Sims presented locomotion on land and water,
and this result has been repeated for many different purposes by many
different researchers.

Lipson and Pollack evolved creatures for locomotion in a
system that allowed the results to be 3-D printed and
activated in the real world~\cite{Lipson2000}.
Creatures composed of rigid segments and linear actuators were
evolved for locomotion in physical simulation.  The body parts
(including joints) could then be 3-D printed, and only the
fitting of actuators required special attention during
assembly.  Notably, these creatures were required to maintain
static stability at all times (i.e., have their center of
gravity always supported by the body).  In this manner, a
consistent transition to the real world was guaranteed, where
dynamics might differ from simulation, but geometry should
not.

Shim and Kim evolved virtual creatures for another type of
locomotion---flight~\cite{shim2003generating}.
Auerbach and Bongard tested the environment's influence on morphology when evolving locomotion~\cite{Auerbach:2012:REM:2330163.2330238}.
Lehman and Stanley used locomoting EVCs as the subject for an
investigation of novelty promotion~\cite{lehman2011evolving}.
Cheney et al.
demonstrated their new encoding for soft-bodied EVCs by
applying them to the locomotion
benchmark~\cite{Cheney:2013:UEE:2463372.2463404}.


\subsection{Phototaxis}

Phototaxis (the ability to move to a light source) was the
most complex behavior demonstrated by Sims.  By testing the
ability to move toward a light target placed at multiple
positions, creatures were developed with a generalized ability
to perform phototaxis.  This remained the most complex EVC
behavior for almost two decades until the work described here.

Pilat and Jacob reproduced the behavioral complexity of Sims'
phototaxis approximately in their 2010
work~\cite{Pilat:2010:EVC:1830483.1830502}, although their
implementation differed in some respects.  While Sims'
photoreceptors were embedded in each body segment and produced
signals relative to the segment's orientation, Pilat and Jacob
used a single sensor for the entire creature, and that
sensor's signals were preprocessed to give one output for
heading to the light and another for the light's elevation
angle.  Also, Pilat and Jacob's creatures had simpler
morphology, allowing only single-degree-of-freedom hinge
joints between segments.  Unlike the control networks of
Sims, with nodes computing a variety of predefined functions,
Pilat and Jacob's creatures used a more conventional
artificial neural network (ANN).
Shim and Kim also achieved a similar result in 2004 with
flying creatures able to follow paths~\cite{shim2004evolving}.
The EVC system presented here demonstrates phototaxis as an
intermediate step on the path to more complex behaviors.

Miconi's work~\cite{miconi2008silicon} is a
particularly interesting case, 
as he is the first to produce a form of real combat between
EVCs, but with respect to behavioral complexity as defined above,
his creatures do not differ significantly from those
of Sims, as their combat can be essentially viewed as target
following with damage assignment layered on top---the target
following leads to collisions, and these collisions 
produce a score interpreted as damage, but no additional
behavioral complexity is required or produced.


\subsection{Task Decomposition}

It is important to note that task-decomposition strategies similar to ESP have been employed in
multiple related fields, but always in conjunction with a fixed morphology.  Selfridge's
pandemonium, Minsky's society
of mind~\cite{minsky1988society}, and Brooks' subsumption
architecture~\cite{brooks1986robust} are prominent examples
from artificial intelligence and robotics.  And in
reinforcement learning and evolutionary computation, work such
as layered learning and hierarchical task
decomposition~\cite{stone2000layered,whiteson:mlj05,doucette2012hierarchical} explores
similar concepts.  In EVCs, however, with the particular challenges of simultaneously evolved morphology and control, no previous system has
demonstrated the use of such an approach to increase
behavioral complexity.


\section{Underlying EVC System}
\label{sec:underlying_EVC_system}

\begin{figure}[ht]
  \setlength{\belowcaptionskip}{-5pt}
  \centering
  \includegraphics[width=0.45\textwidth]{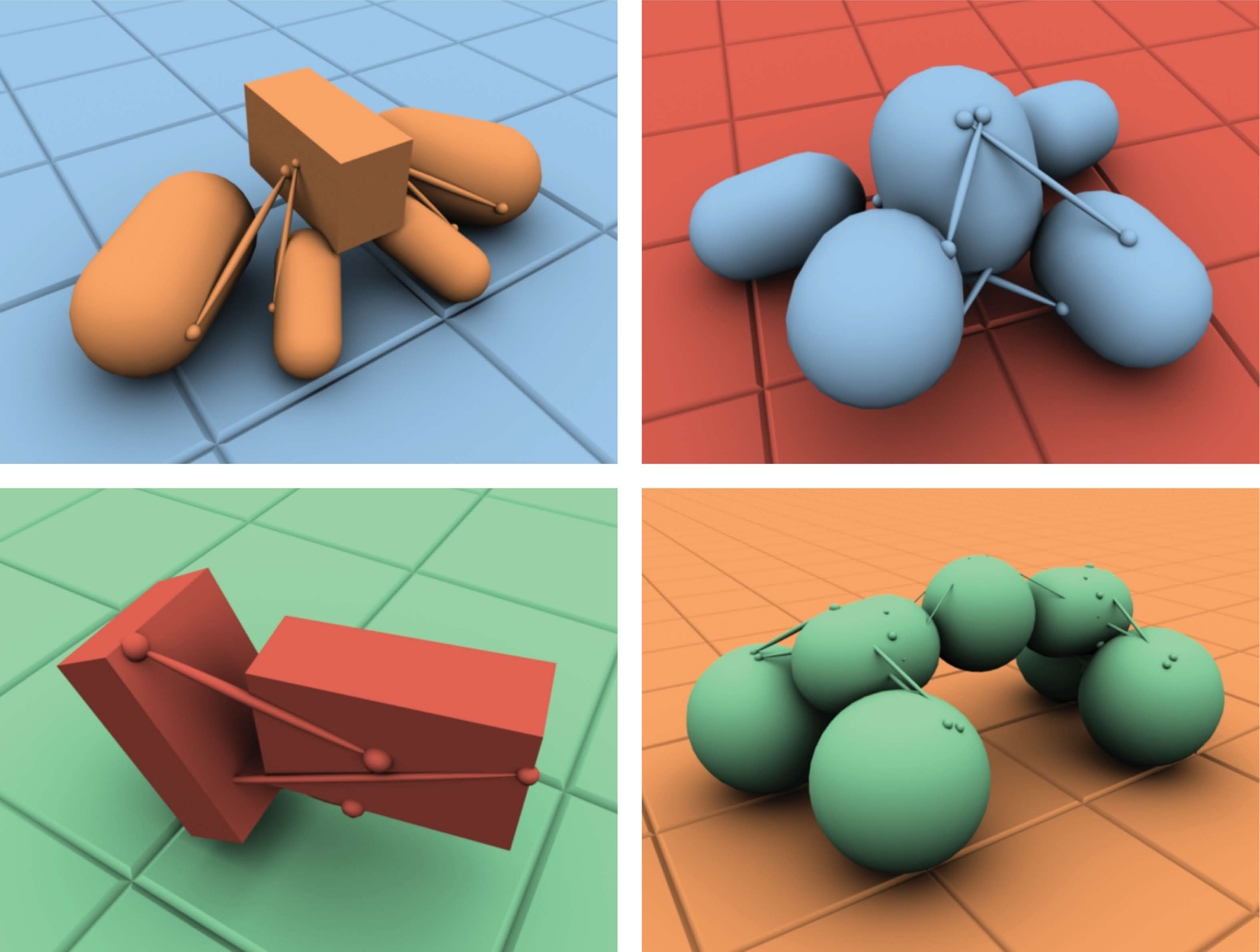}
  \caption{Typical results from the underlying basic EVC system.  These 
  examples were all evolved to complete a forward locomotion
  task---a common baseline result for EVCs.}
  \label{fig:basic_evc}
\end{figure}

The underlying EVC system described here is largely derived from 
the work of Karl Sims~\cite{Sims:1994:EVC:192161.192167}. This
section briefly sets out the  
components of this system, which---while not the primary focus
of this paper---are nevertheless fundamental to its
comprehension.  A representative sample of results is shown in
Figure~\ref{fig:basic_evc}.

\subsection{Evolutionary Algorithm}

The specifics of
the evolutionary algorithm are conventional, making use of
elitism, fitness-proportionate selection, and rank
selection~\cite{Mitchell:1998:IGA:522098}.  In addition, the
most challenging tasks employ some degree of
shaping~\cite{skinner1938behavior}.  Fitness
is evaluated in a physically simulated virtual environment
implemented with NVIDIA PhysX.

\subsection{Morphology}

\begin{figure}[t]
  \setlength{\belowcaptionskip}{-5pt}
  \centering
  \begin{subfigure}{1.6in}
    \centering
    \includegraphics[width=0.5625in, height=0.5625in, keepaspectratio=true]{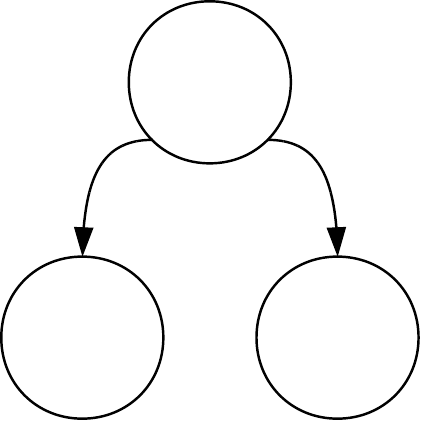}
    \hspace{0.05in}
    \includegraphics[width=0.75in]{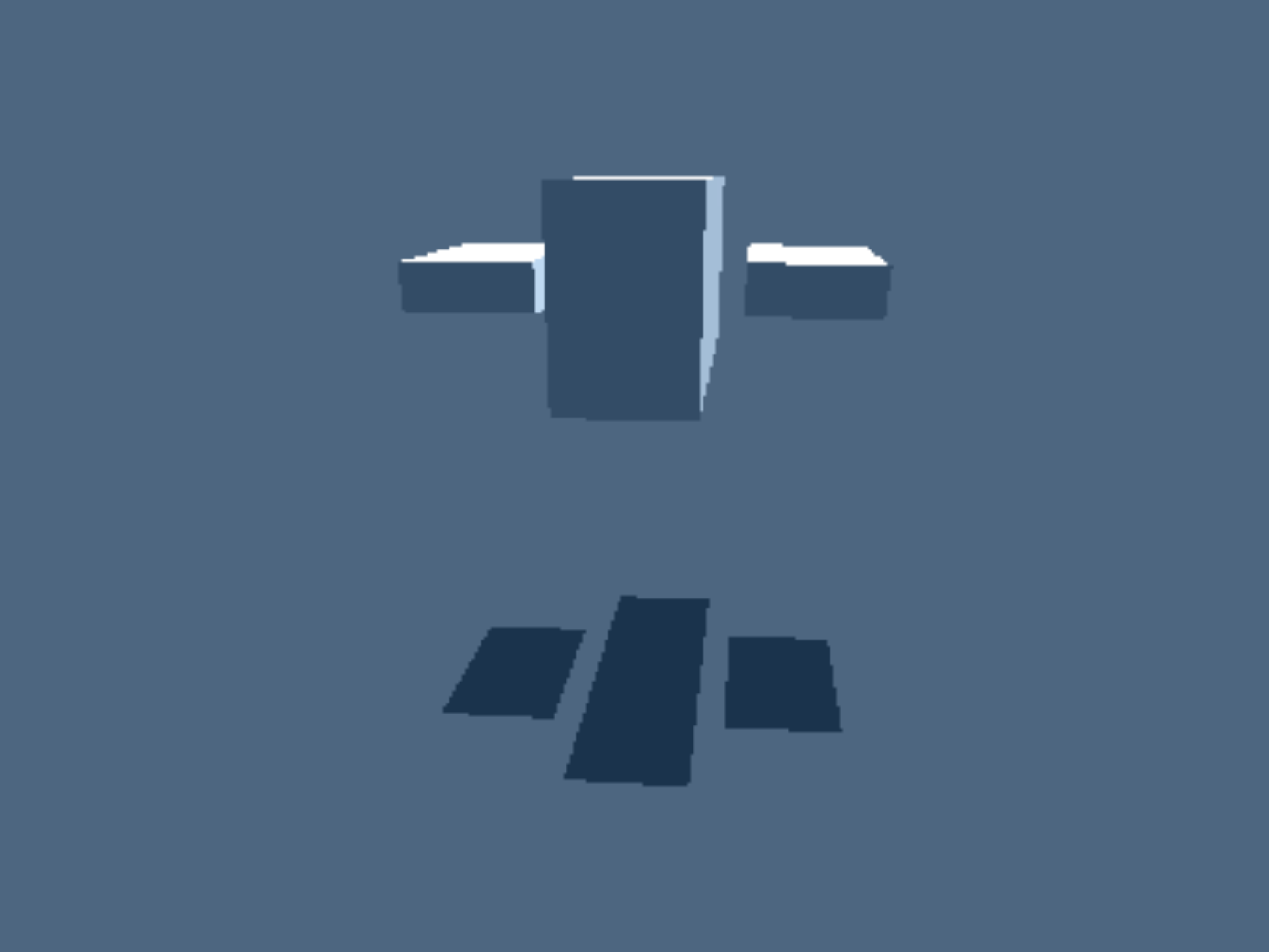}
    \caption{Simple topology.}
  \end{subfigure}
  \begin{subfigure}{1.6in}
    \centering
    \includegraphics[width=0.5625in, height=0.5625in, keepaspectratio=true]{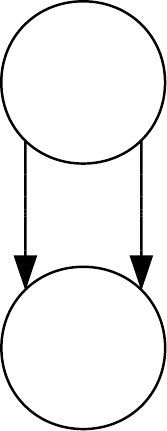}
    \hspace{0.05in}
    \includegraphics[width=0.75in]{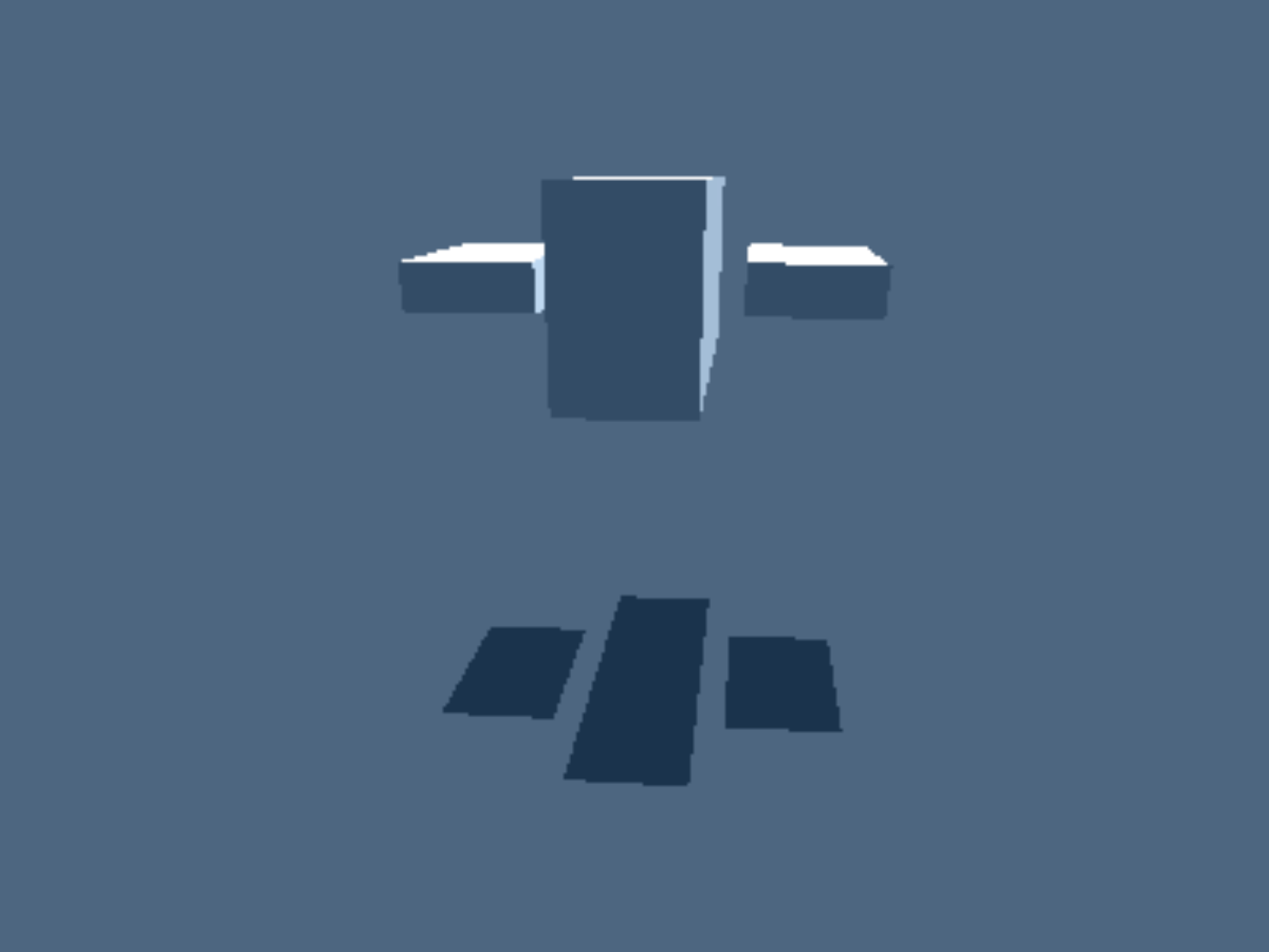}
    \caption{Multiple edges for repeated substructures.}
  \end{subfigure}
  \vspace{0.1in}

  \begin{subfigure}{1.6in}
    \centering
    \includegraphics[width=0.35in, height=0.35in, keepaspectratio=true]{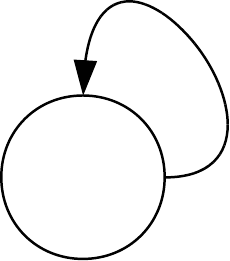}
    \hspace{0.05in}
    \includegraphics[width=0.75in]{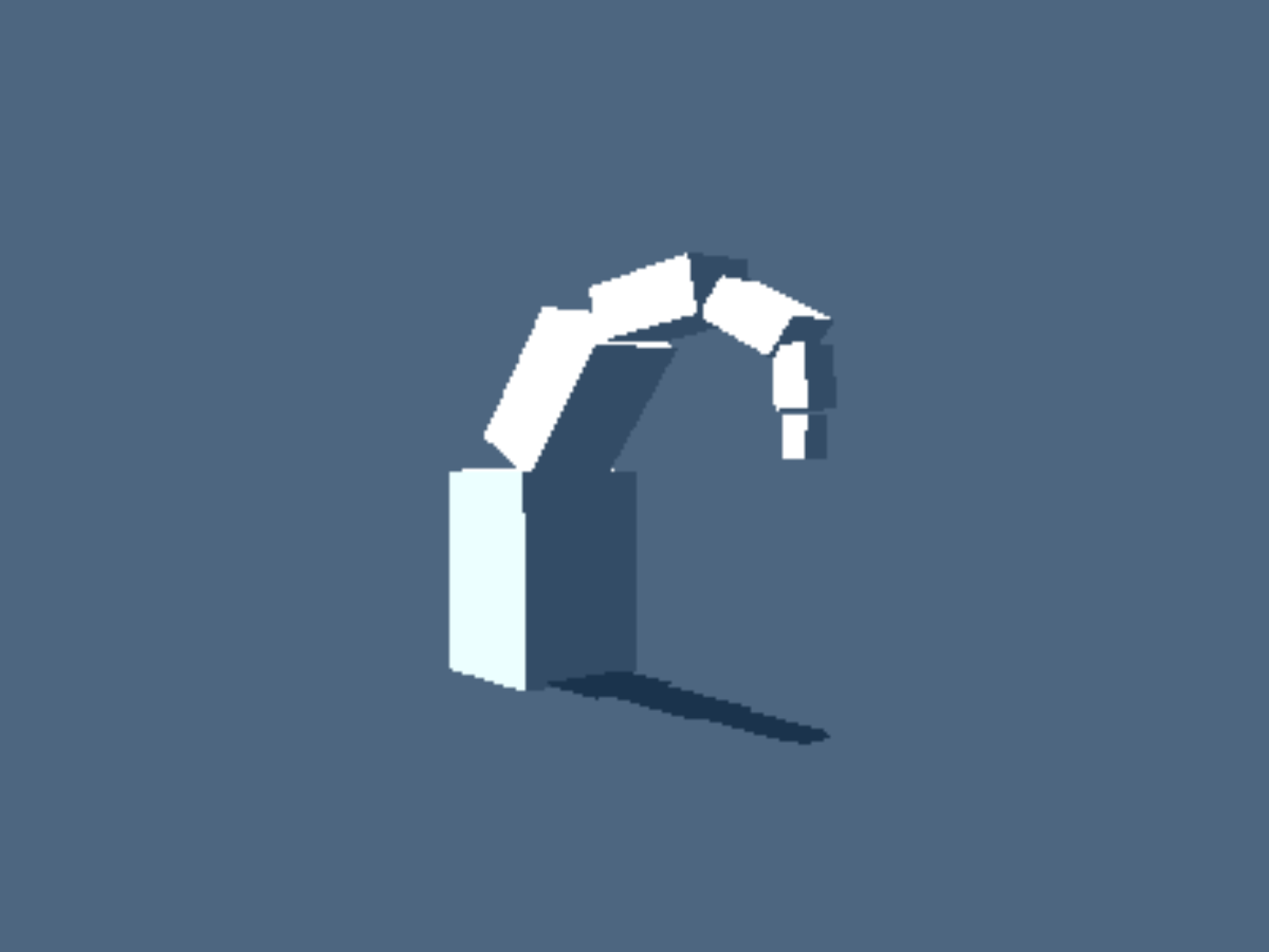}
    \caption{Reflexive edge for recursive structure.}
  \end{subfigure}
  \begin{subfigure}{1.65in}
    \centering
    \includegraphics[width=0.75in, height=0.5625in, keepaspectratio=true]{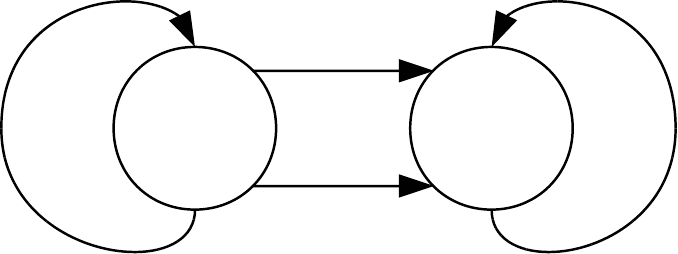}
    \hspace{0.05in}
    \includegraphics[width=0.75in]{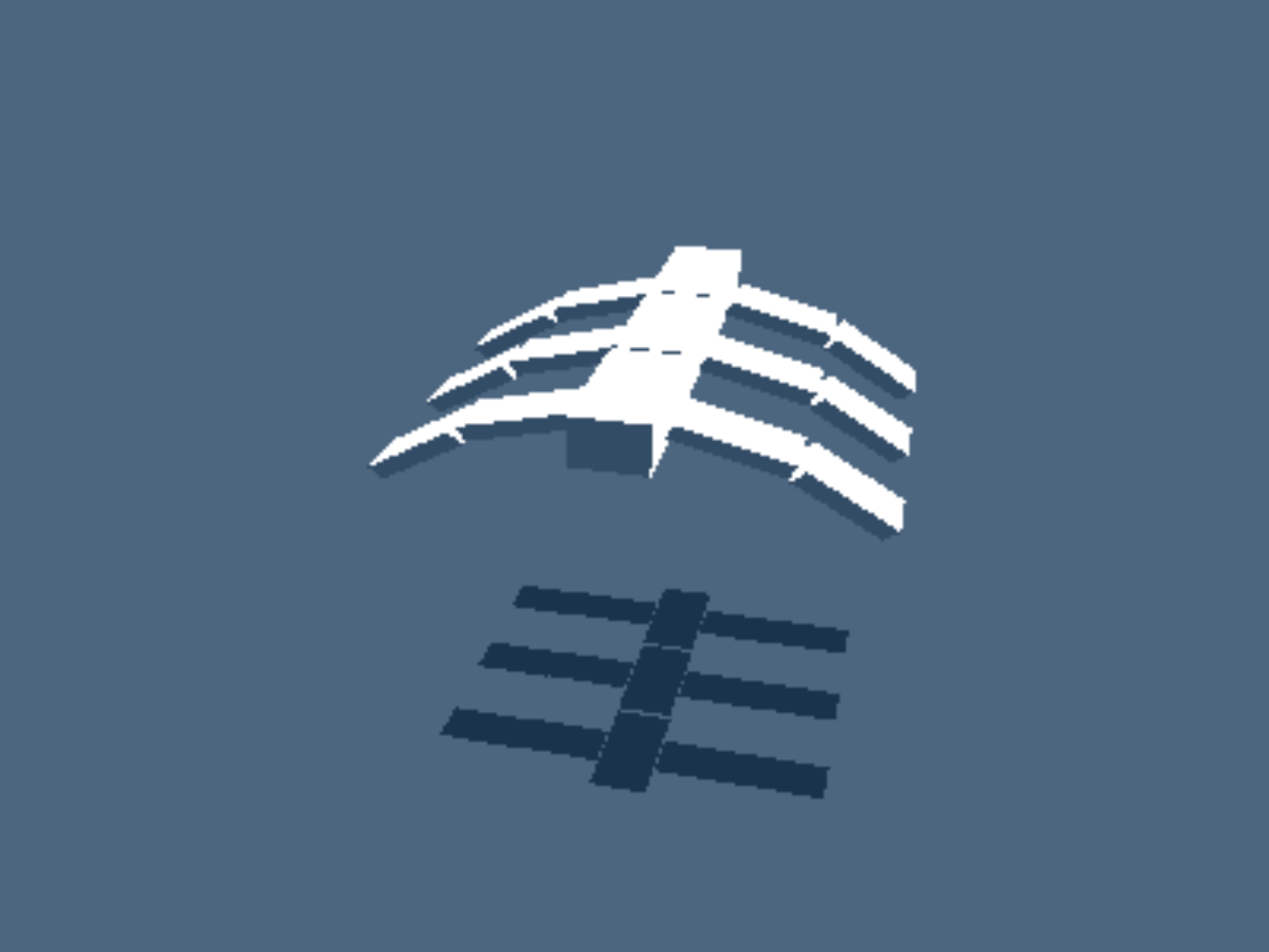}
    \caption{Multiple and reflexive edges together.}
  \end{subfigure}
  \vspace{0.1in}

  \caption{
    Hand-designed genotype/phenotype pairs (as
    in~\protect\cite{Sims:1994:EVC:192161.192167}) demonstrate
    the encoding power inherited from Sims' original EVC
    system.}
  \label{fig:encoding}
\end{figure}

As in Sims' original work~\cite{Sims:1994:EVC:192161.192167},
creature morphology is described by 
a graph-based genotype, with graph nodes
representing body segments, and graph edges representing
joints between segments.  By starting at the root and
traversing the graph's edges, the phenotype is expressed. 
Reflexive edges as well as multiple edges between the same
node pair are allowed, making it possible to easily define
recursive and repeated body substructures, as illustrated in
Figure~\ref{fig:encoding}.  In addition, as in 
Sims' work, reflection of body parts as well as body symmetry
are made easily accessible to evolution.  In this
implementation, all PhysX primitives are made available for
use as body segments: boxes, spheres, and capsules. 
Joints between segments may be of most of the types offered by
PhysX, specifically: fixed, revolute, spherical, prismatic,
and cylindrical.  In contrast to the typical technique of
separately evolving explicit joint limits, most limitations on
joint movement in this system are provided implicitly by
creature structure through natural collisions between adjacent
segments.

\subsection{Control}

Again in a manner very similar to that of Sims, creature
control is provided by a brain composed of a set of nodes
connected by wires (as in Figure~\ref{fig:encapsulation_before}).
Nodes receive varying numbers of input wires, and use their
inputs to compute an output value (always in the range [0,1])
which may be sent to other wires.  Signals originate from
sensors in the body as well as certain types of internal brain
nodes, travel through the network of internal nodes and wires,
and ultimately control the operation of actuators (muscles) in
the physically simulated body.  For each step of physical
simulation, control signals move one step through the 
brain.

In addition to special node types for
muscles and photoreceptors (described below) and one special
type used in encapsulation (see Section~\ref{sec:encapsulation}), the following node
types are allowed: sinusoidal, complement, constant, scale,
multiply, divide, sum, difference, derivative, threshold,
switch, delay, and absolute difference.

\subsection{Photoreceptors}

\begin{figure}[t]
  \setlength{\belowcaptionskip}{-5pt}
  \centering
  \includegraphics[width=2.5in]{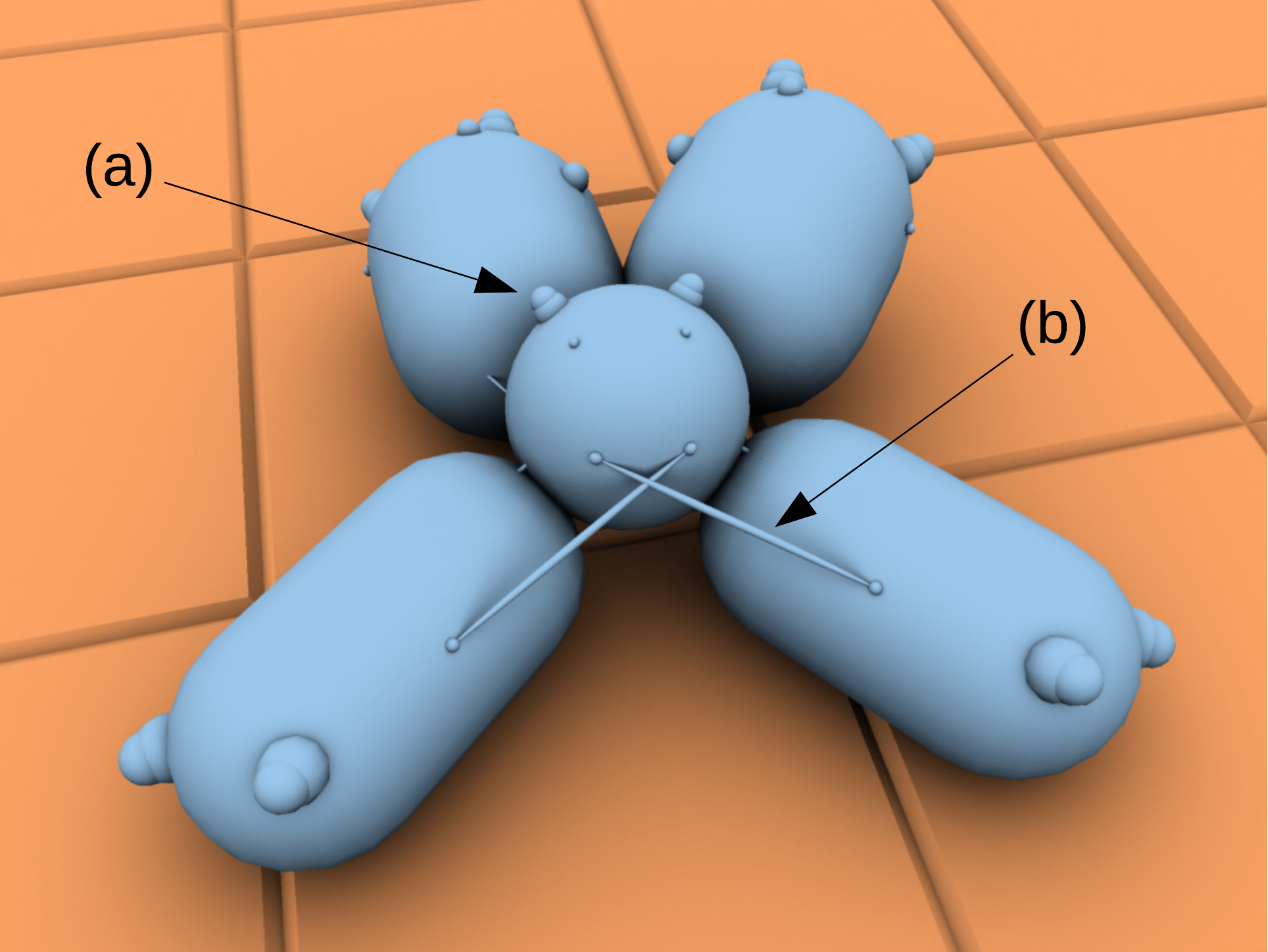}
  \caption{Photoreceptors (a) and muscles (b) bring sensing
    and actuation to creatures in the underlying EVC system.  For
    both, function depends upon placement, so creature form
    develops meaningfully as capabilities are evolved.}
\label{fig:photoreceptors_and_muscles}
\end{figure}

For tasks involving light sensing, creatures are allowed to
develop simple photoreceptors
((a) in Figure~\ref{fig:photoreceptors_and_muscles}), defined only
by a direction 
from the center of their parent segment.  This direction
indicates a location on the creature's surface as well as an
orientation for the receptor.  The signal produced by the
receptor is determined by light strength, distance, occlusion,
and the difference between the direction to the light and the
sensor's orientation.  Multiple lights are allowed.  For
each photoreceptor in the body, a corresponding brain node is
added which makes the receptor's output signal available to the
rest of the brain.

\subsection{Muscles}

In a break with traditional EVC systems, which typically use
forces exerted directly at joints, this system uses simulated
muscles as actuators.  Each muscle
((b) in Figure~\ref{fig:photoreceptors_and_muscles}) is defined by two
attachment points on adjacent segments, along with a maximum
strength value.  In simulation, the muscle is implemented as a
spring, with muscle activation modifying the spring constant.
In addition to acting as an effector, each muscle also produces a
proprioceptive feedback signal based on its current length.
For each muscle, one node is added to the brain which accepts
an input to set the muscle's activation, and another node is added
which makes the muscle's proprioceptive output signal
available to the rest of the brain.  Muscle drives bring the
following potential benefits to EVCs: flexibility (they can be
used even on creatures without joints), efficiency (effectors
need only exist where useful, not at every degree of freedom
of every joint), and aesthetic appeal (by tapping into the human
affinity for elegant, functional body structure).


\section{Fast ESP}
\label{sec:ESP}

Being simpler than General ESP, the Fast ESP method will be described first.  Note that with the exception of limiting morphological changes, as described in Section~\ref{sec:fast_esp_morphological_limitations}, all elements of the method described here are also a part of General ESP (Section~\ref{sec:extended_ESP}).

The Fast ESP method~\cite{Lessin:2013:OBC:2463372.2463411} consists of three elements added to
the underlying EVC system: a syllabus, 
encapsulation, and pandemonium.  In the beginning of this section, each of
these components is described in detail.  The section ends with a description of the morphological limitations specific to Fast ESP.

\subsection{Syllabus}

While it is certainly possible for human students to learn a
complicated topic independently, their development is
typically faster and surer with the benefit of an
expert-designed syllabus.  The syllabus acts as a sequence of
landmarks through the space of possible solutions, decomposing
the larger learning task into a succession of more manageable
steps between these waypoints.

In the ESP system, the \emph{syllabus} consists of an
ordered sequence of fitness goals used to reach the ultimate,
larger goal.  This collection of intermediate goals (each one
defined by a fitness function) is designed by a human expert
with the aim of making attainable goals more reliably
learnable, and bringing previously unattained goals within
reach.

For example, assume that you want to evolve a virtual creature
with some of the behavioral complexity demonstrated in an
internet cat video.  Rather than simply drifting smoothly toward
a target, this creature might run to the target, then strike it, and
perhaps even run away if the target is perceived as
threatening.  Without a syllabus, a single fitness test
evaluating all of these skills might be constructed, but
evolutionary progress would be unlikely.

Consider, instead, how this complex behavioral goal could be
broken down into an 
ordered sequence of smaller learning tasks.  The clearly
achievable goal of locomotion will be the first 
target.  The ability to turn left and the ability to turn
right are of a similarly manageable difficulty, and will
be attempted next.  Then, with left and right turns mastered,
and the ability to develop photoreceptors, it would seem
relatively straightforward to maintain orientation toward a
light source.  And with the ability to face a light and
the ability to move forward, navigating to that light might be
a similarly achievable goal.  Proceeding in this manner, a
knowledgeable human designer might produce the following
sequence of subskills to be learned, in which each subskill is
probably attainable with basic EVC methods, and in which
earlier subskills serve to make later skills easier to learn:

\begin{figure}[t]
  \setlength{\belowcaptionskip}{-5pt}
  \centering
  \includegraphics[height=3.0in]{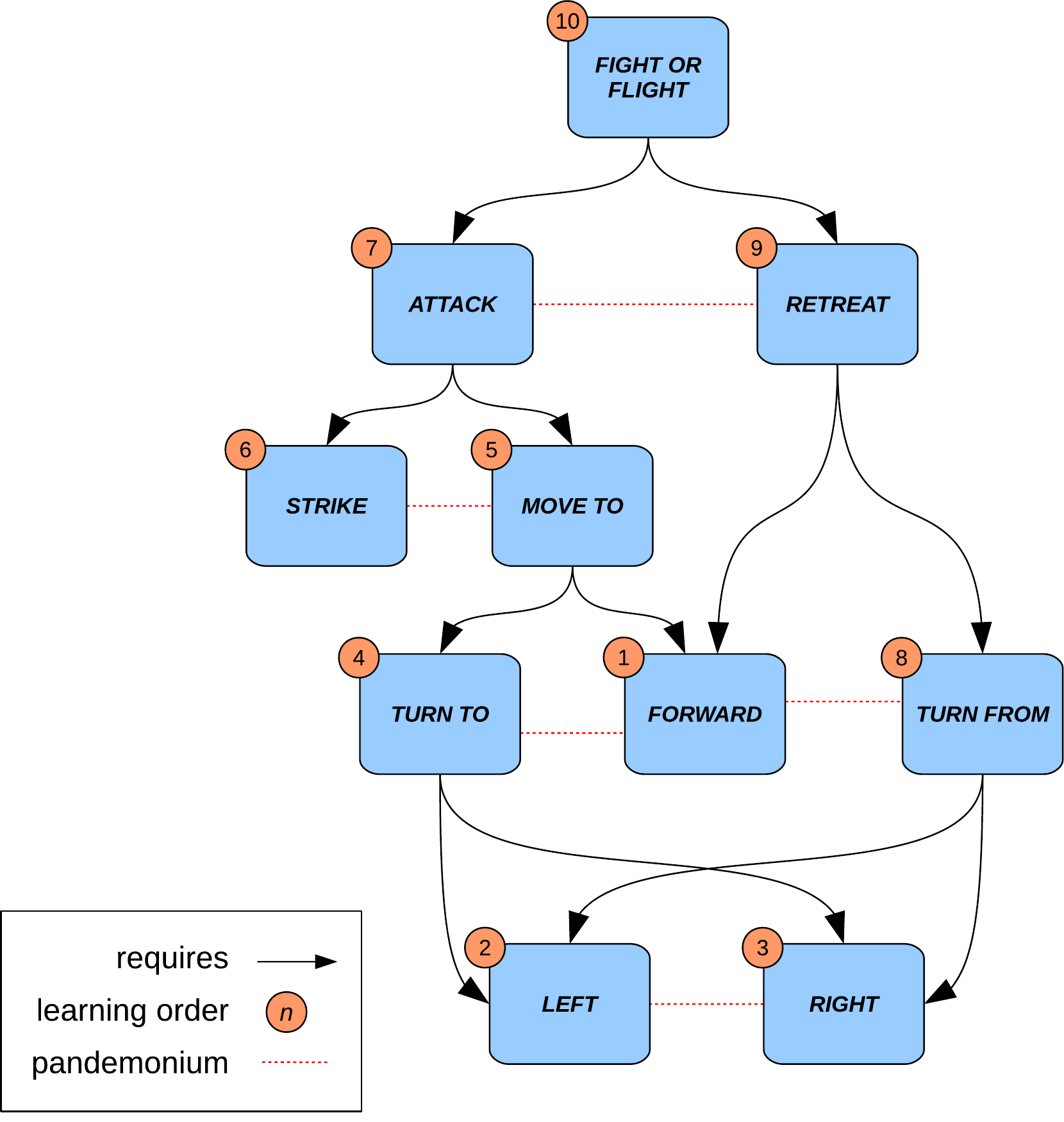}
  \caption{An example syllabus as a graph.  In this depiction, 
graph nodes represent individual subskills to be learned,
directed edges indicate dependency between subskills, and the
numbering indicates a proposed learning order which satisfies
the dependency requirements.  Pandemonium relationships are
indicated by dashed red lines.}
\label{fig:syllabus_graph}
\end{figure}

\begin{enumerate}
\item \textsc{forward locomotion}
\item \textsc{left turn}
\item \textsc{right turn}
\item \textsc{turn to light} (using \textsc{left turn} and
  \textsc{right turn})
\item \textsc{move to light} (using \textsc{turn to light} and
  \textsc{forward locomotion})
\item \textsc{strike}
\item \textsc{attack light} (using \textsc{move to light} and
  \textsc{strike})
\item \textsc{turn from light} (using \textsc{left turn} and
  \textsc{right turn})
\item \textsc{retreat from light} (using \textsc{turn from
  light} and \textsc{forward locomotion})
\item \textsc{fight or flight} (switching between
  \textsc{attack light} and \textsc{retreat from light} based
  on external circumstances)
\end{enumerate}

This information may be conveniently summarized in a graph,
encompassing subskills to be learned, dependency between
subskills, learning order, and pandemonium
(Section~\ref{sec:pandemonium}), as seen in
Figure~\ref{fig:syllabus_graph}.

At this point, using high-level human knowledge, a previously
impractical learning task has been broken into a 
sequence of potentially attainable subgoals.  But how can a
single evolving creature learn new skills while retaining and
making use of the ones it already has?

\subsection{Encapsulation}
\label{sec:encapsulation}

\begin{figure}[t]
  \setlength{\belowcaptionskip}{-5pt}
  \centering
  \begin{subfigure}[c]{1.6in}
    \centering
    \includegraphics[width=1.6in]{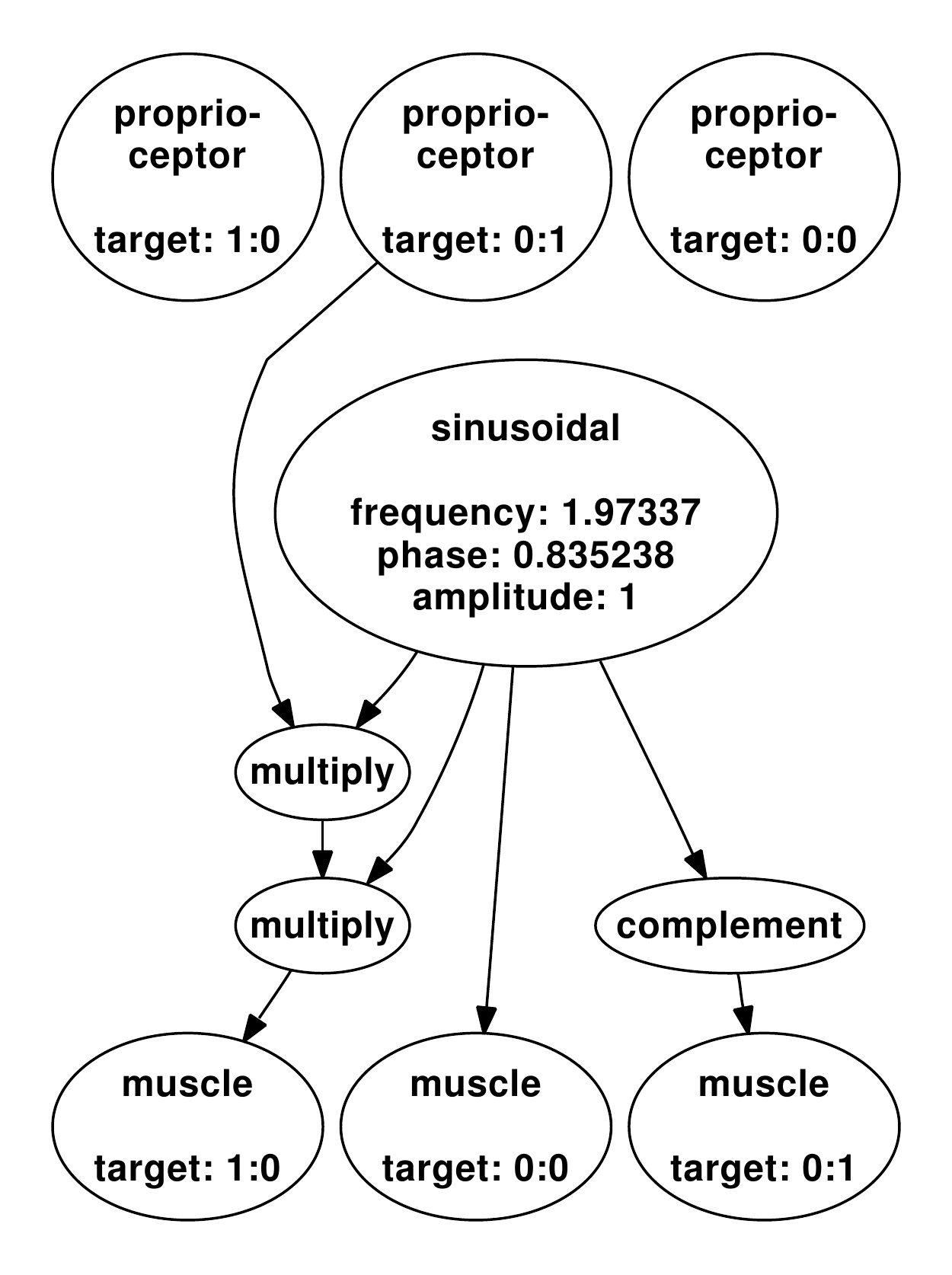}
    \caption{Before encapsulation.}
    \label{fig:encapsulation_before}
  \end{subfigure}
  \begin{subfigure}[c]{1.4in}
    \centering
    \includegraphics[width=1.4in]{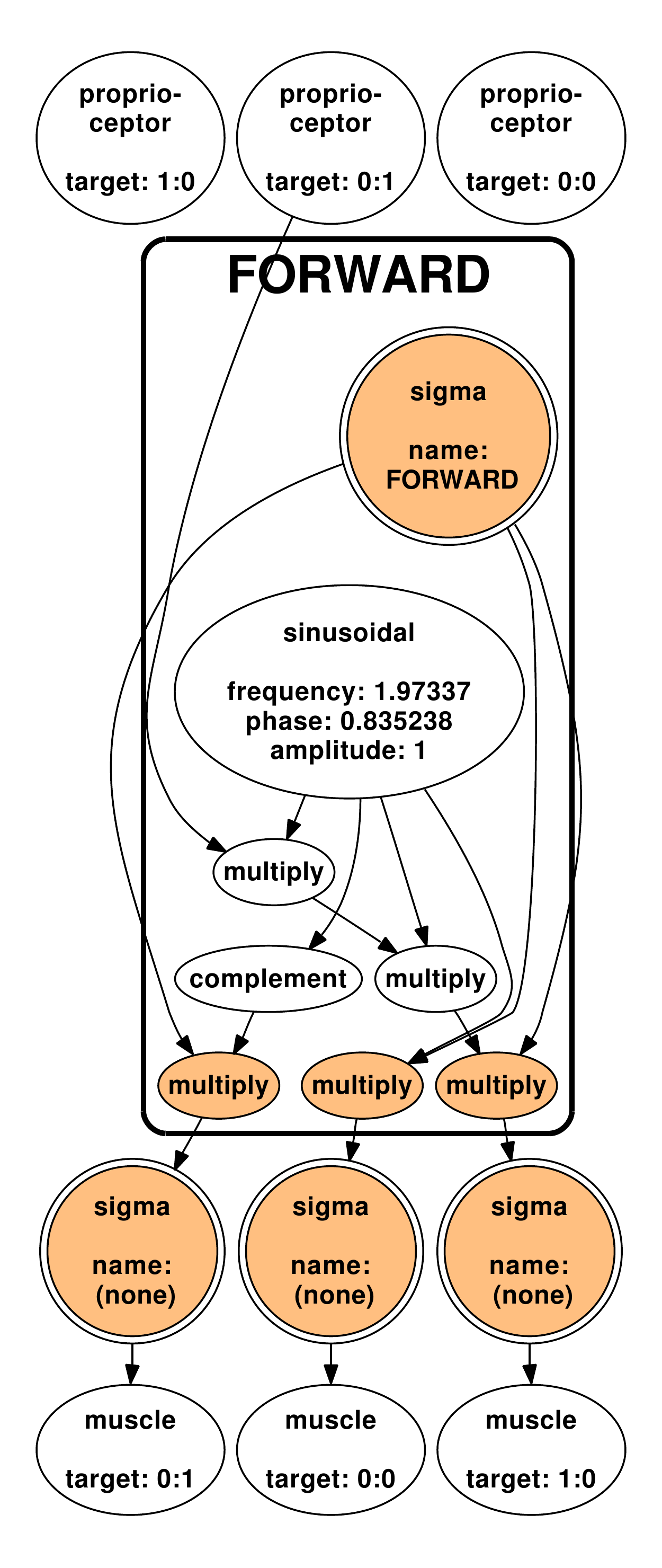}
    \caption{After encapsulation (with new nodes shaded).}
    \label{fig:encapsulation_after}
  \end{subfigure}
  \caption{The automated encapsulation of an evolved
    skill---in this case, forward locomotion---ensures that it
    will persist throughout future evolution, while also
    allowing it to be easily activated as a unit by future
    skills.}
  \label{fig:encapsulation}
\end{figure}

The second important element of the ESP system is a
mechanism to \emph{encapsulate} previously learned skills.
This accomplishes two important goals: It ensures that
previously learned skills (and the body components they rely
on) are preserved, and it makes these skills easily accessible
to future evolutionary development.  Both of these goals are
achieved through the automated encapsulation process
illustrated in Figure~\ref{fig:encapsulation}.

Figure~\ref{fig:encapsulation_before} depicts a brain
evolved for forward locomotion, and
Figure~\ref{fig:encapsulation_after} shows the result of
encapsulation.  Note the following aspects of this new brain.
The nodes that compute the old skill have been preserved and
locked (meaning that they have been marked so as to disallow
any changes by future evolution).  Also, a new \emph{multiply}
node has 
been inserted into every output wire leaving the encapsulated
skill.  The internals of the skill will continue to function
as before, always trying to perform their forward locomotion
task, but now, a second signal sent to each new
multiply node will modify those outgoing forward-locomotion
control signals, scaling them by a number in [0,1].  
Finally, a single controlling node (called a 
\emph{sigma node} for its function as a summation of zero or
more inputs) is added, which sends its output to all of the
new multiply nodes.  So for each signal $s_i$ leaving
a node in the \textsc{forward locomotion} skill (such as the
\emph{complement} node), the new signal after encapsulation
($s'_i$) is computed as $s'_i = \sigma s_i$ where $\sigma$ is
the output of the controlling sigma node.

Now, with encapsulation complete, the entire forward locomotion
skill can be activated and deactivated as a unit by
using the controlling sigma node just as if it were a
single muscle in the creature's body. (Incidentally, note that
this brain's actual muscle nodes have been hidden behind
additional sigma nodes to allow future evolution to share
control over them when appropriate.)
As progress through the syllabus continues and the next skill
after \textsc{forward locomotion} is evolved, its newly added
nodes will be the only ones in the brain that are not already
locked, and will therefore be easily identifiable when it is
their turn to be encapsulated.

At this point, we have seen a system in which a complex skill
can be broken into smaller subskills, and those subskills can
be cumulatively acquired, but a potential problem still
remains: How will competing signals from the multiple
sub-brains within a single creature be resolved?

\subsection{Pandemonium}
\label{sec:pandemonium}

Consider the following example based on the syllabus graph of
Figure~\ref{fig:syllabus_graph}.  A creature evolved through
this syllabus will ultimately have parts of its brain devoted
to both left and right turns.  But it seems unlikely that both
of these abilities should ever be used at the same 
time.  So the syllabus designer might place the left and right-turn skills in a 
\emph{pandemonium} relationship with each other, meaning that
whichever one is most active at any given moment will be
allowed to send its output at full strength, and the other
will have its output entirely suppressed.  Under a system like
this, sub-brains within the creature can compete for overall
control, with little risk of sabotaging the usefulness of the
entire brain.  In Figure~\ref{fig:syllabus_graph}, a full set
of pandemonium relationships is indicated by red dashed lines
between subskill nodes.

With this final component of the ESP system described, it is
now possible to consider a full example, in which previously
achieved levels of behavioral complexity are first matched,
then exceeded.

\subsection{Morphological Limitations in Fast ESP}
\label{sec:fast_esp_morphological_limitations}

In order to obviate retesting of previously learned skills, Fast ESP limits morphological changes afer the first skill in the syllabus is complete.  Because of this, only the new skill being learned must be evaluated during evolution, leading to an approximately linear growth in computation time with respect to the number of skills learned.  Body changes with no significant impact on existing skill function---the addition of eyes and muscles---are permitted throughout development, but changes which might invalidate existing control abilities--those to the skeleton segments and joints---are prohibited after the acquisition of the first skill is complete.


\section{Fast ESP Results}
\label{sec:ESP_results}

\begin{figure}[t]
  \setlength{\belowcaptionskip}{-5pt}
  \centering
  \begin{subfigure}{0.45\textwidth}
    \centering
    \includegraphics[width=\textwidth]{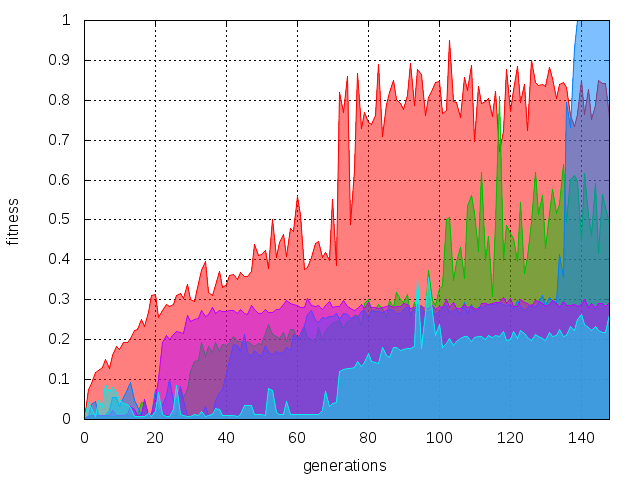}
    \caption{}
  \end{subfigure}
  \begin{subfigure}{0.45\textwidth}
    \centering
    \includegraphics[width=\textwidth]{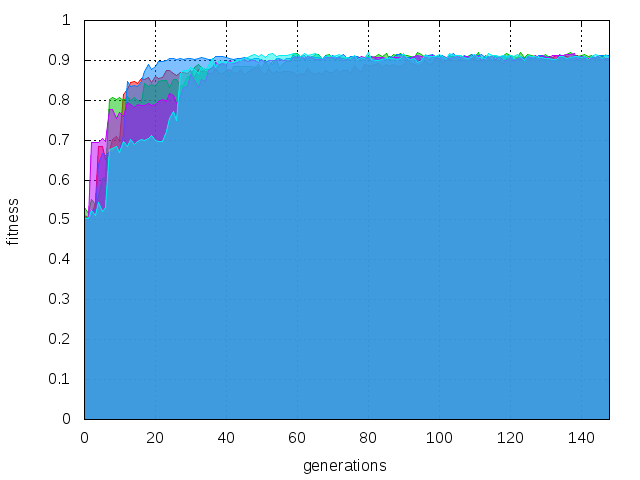}
    \caption{}
  \end{subfigure}
  \caption{Better fitness development with ESP.  Fitness graphs for a skill that controls the body directly and for a skill constructed hierarchically from existing skills.
  (a) Fitness graphs for all five runs of the \textsc{left turn} skill.  Since this behavior must develop full morphological control from scratch, progress may be irregular and inconsistent.
  (b) Fitness graphs for all five runs of the \textsc{turn to light} skill.  By taking advantage of ESP's ability to re-use existing encapsulated behaviors, creatures can often acquire such abilities quickly and consistently.
  These plots demonstrate ESP's ability to make complex skills easier to acquire.}
  \label{fig:fast_esp_fitness_graphs}
\end{figure}

The primary result of this paper is an application of the ESP
method, using the syllabus of Figure~\ref{fig:syllabus_graph},
to evolve a virtual creature through a sequence of ten
learning tasks, the first five of which approximately match
the previously demonstrated behavioral-complexity limit for
EVCs, and the second five of which approximately double it.
In this section, this is demonstrated with the Fast ESP implementation.
(These results are best viewed in the first accompanying
video\footnote{http://youtu.be/dRLNnJlT8rY}.)


\subsection*{\footnotesize \textbf{Skill 1: FORWARD LOCOMOTION}}

\begin{figure}[ht!]
  \setlength{\belowcaptionskip}{-10pt}
  \includegraphics[height=1.0in]{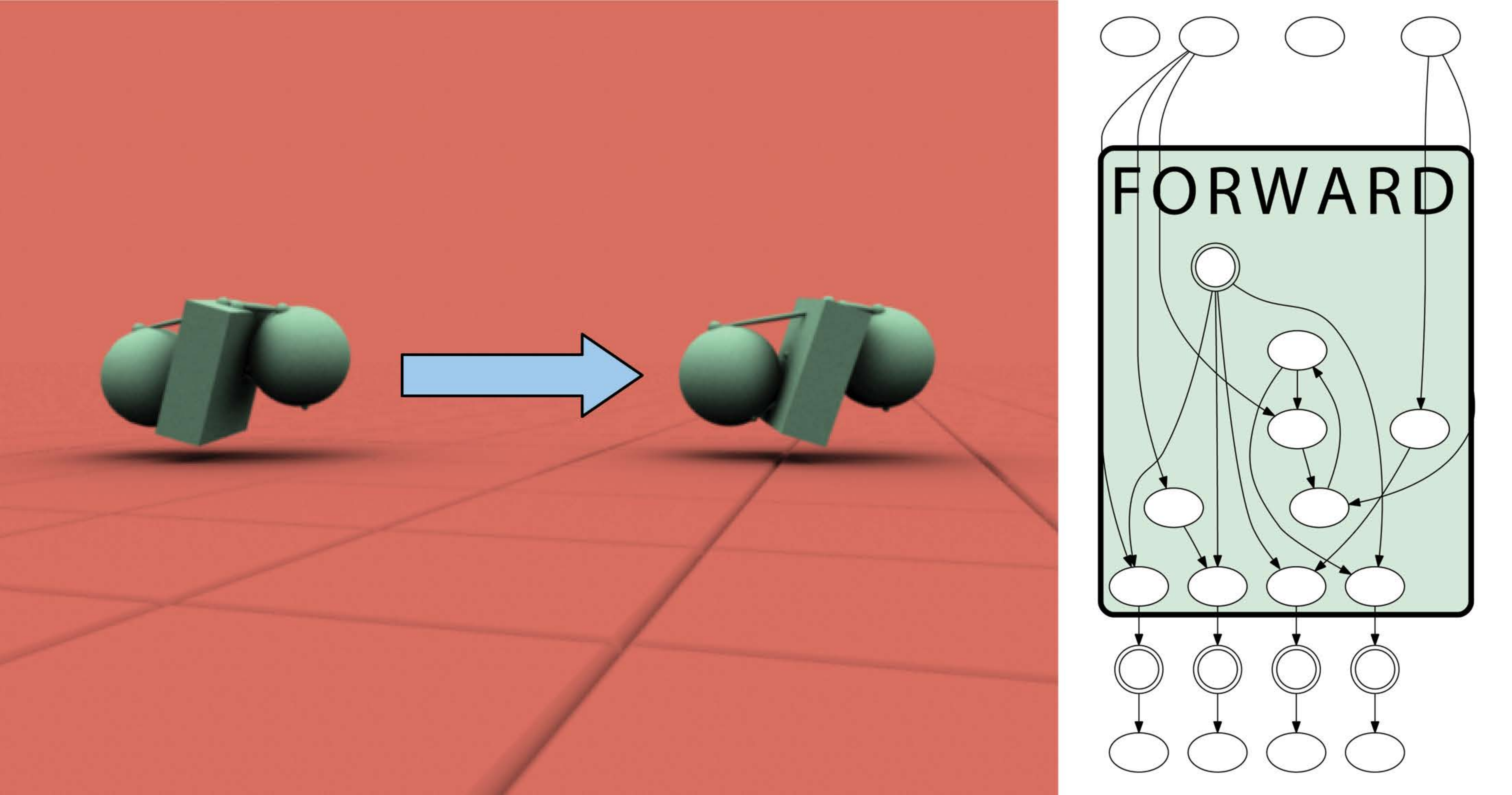}
  \caption{\textsc{forward locomotion} encapsulated.}
  \label{fig:results_01}
\end{figure}

A \textsc{forward locomotion} result from the basic EVC system
has been chosen, and its control abilities have been
encapsulated, as shown in Figure~\ref{fig:results_01}.  This
creature was evolved through traditional EVC techniques,
including the use of shaping, with the ultimate
fitness being defined by the interleaving of an efficiency
score into a discretized score for speed.  Specifically, if
$s$ is the creature's speed, $s_{max}$ is the maximum speed,
$\sigma$ is the discretization step, and $\epsilon$ is a
measure of the creature's efficiency (in [0, 1]), the combined
fitness is defined as
\[
\frac{\sigma (\lfloor \frac{s}{\sigma} \rfloor + \epsilon) }
{s_{max}}.
\]
This is intended to ensure that speed is the primary factor in
fitness, but increased efficiency (while maintaining
approximate speed) is also rewarded.

At this point, the creature has developed the rigid body
segments, muscles, and control system it needs for successful
locomotion and, as a part of the Fast ESP algorithm, these elements will be preserved as evolution
continues.

\subsection*{\footnotesize \textbf{Skill 2: LEFT TURN}}

\begin{figure}[ht!]
\setlength{\belowcaptionskip}{-10pt}
  \includegraphics[height=1.0in]{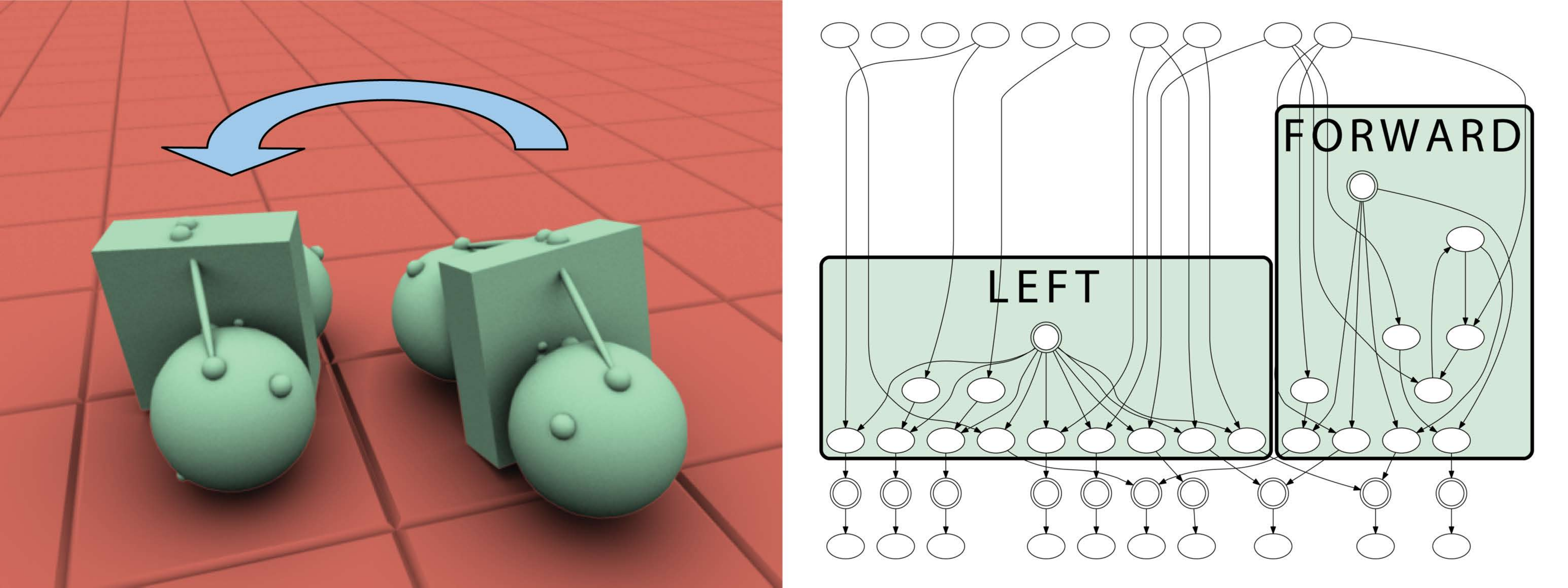}
  \caption{\textsc{left turn} added.}
  \label{fig:results_02}
\end{figure}

With the \textsc{locomotion} skill preserved, a new run of
evolution begins, this time with the fitness function
rewarding the ability to rotate counterclockwise while largely
maintaining position.  The addition of new muscles is allowed
during this process.  The resulting completed skill is shown
(after encapsulation) in Figure~\ref{fig:results_02}.

\subsection*{\footnotesize \textbf{Skill 3: RIGHT TURN}}

\begin{figure}[ht!]
\setlength{\belowcaptionskip}{-10pt}
  \includegraphics[height=1.0in]{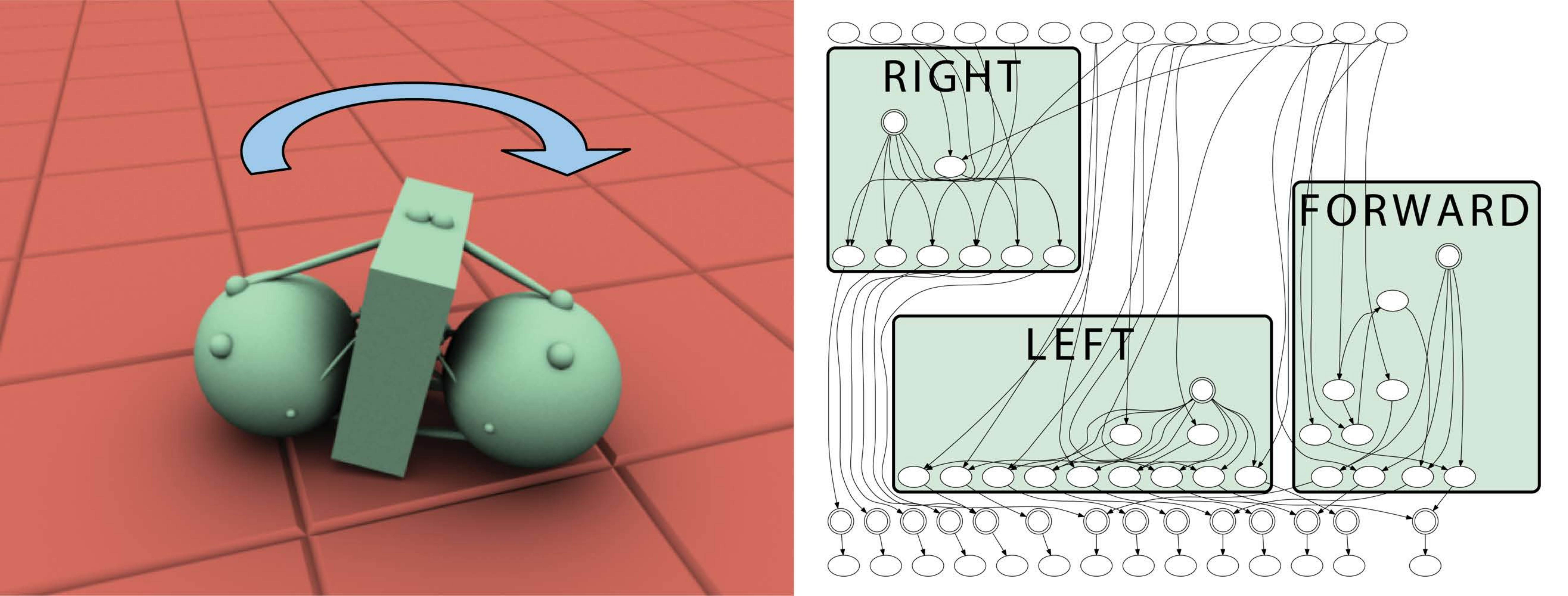}
  \caption{\textsc{right turn} added.}
  \label{fig:results_03}
\end{figure}

With the first two skills preserved, a clockwise turn
is evolved in the same way as the counterclockwise turn, and
the result is encapsulated (Figure~\ref{fig:results_03}).  At
this point, the creature has all of the low-level skills that
it will need to reach any point on the ground, with the
majority of future skills relying ultimately on reapplications
of \textsc{forward locomotion}, \textsc{left turn}, and
\textsc{right turn}.

\newpage
\subsection*{\footnotesize \textbf{Skill 4: TURN TO LIGHT}}

\begin{figure}[ht!]
\setlength{\belowcaptionskip}{-10pt}
  \includegraphics[height=1.0in]{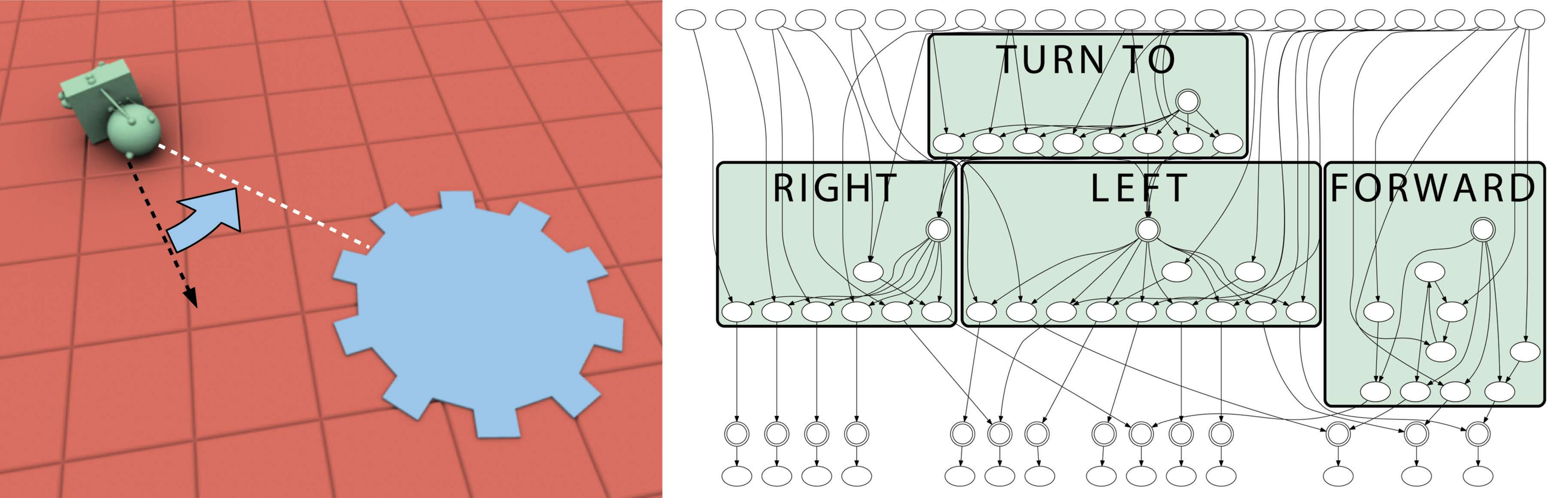}
  \caption{\textsc{turn to light} has been added, which keeps
    the locomotion direction (black dashed arrow) oriented
    toward a target.}
  \label{fig:results_04}
\end{figure}

At this point, the creature is allowed to develop
photoreceptors, while being tested on its ability to orient
itself to a target (which is perceived as a point light
source) using the previously encapsulated \textsc{left turn}
and \textsc{right turn} skills.
The fitness evaluation is an
average over four runs, each with a fixed light source at a
different heading from the creature.
Figure~\ref{fig:results_04} shows the completed and
encapsulated result, which is able to consistently aim its
locomotion direction at a user-controlled target.

Note that skills such as this---which take advantage of ESP's ability to reuse previously encapsulated abilites---can for some tasks produce results very quickly and consistently.  This is in contrast to skills such as \textsc{left turn}, which must solve the potentially much harder problem of full morphological control from scratch.  This contrast is illustrated in the fitness graphs for the two skills, as seen in Figure~\ref{fig:fast_esp_fitness_graphs}.

\subsection*{\footnotesize \textbf{Skill 5: MOVE TO LIGHT (Matches the Previous State of the Art)}}

\begin{figure}[ht!]
\setlength{\belowcaptionskip}{-10pt}
  \includegraphics[height=1.0in]{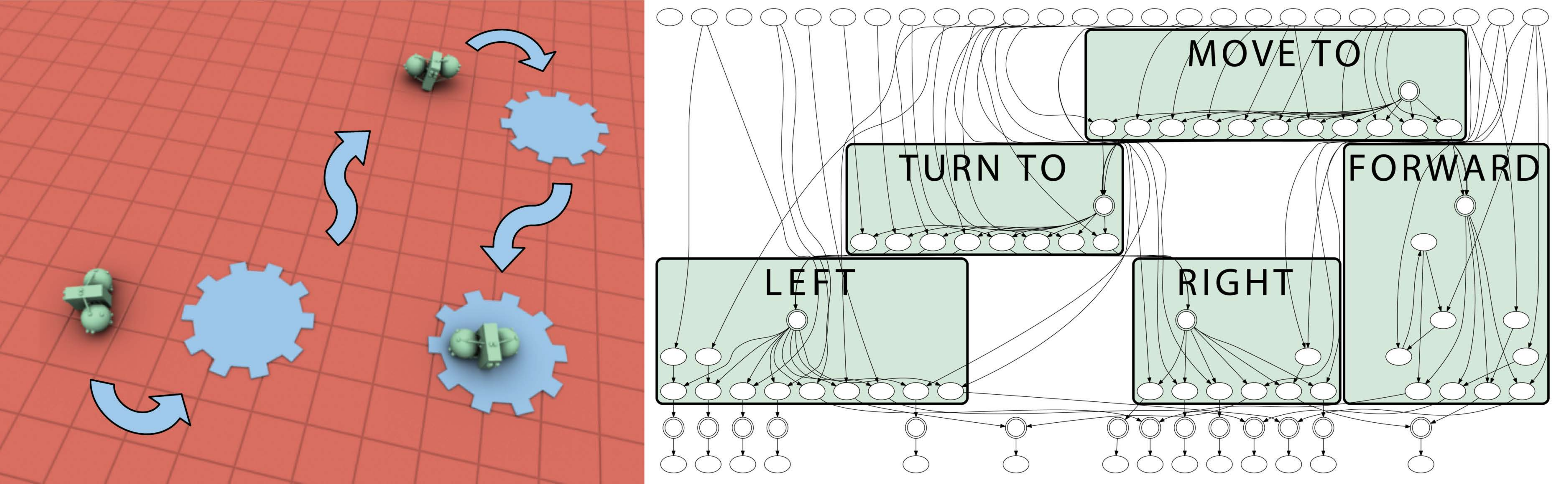}
  \caption{\textsc{move to light} has been added, allowing the
    creature to follow a target along a complex path, catching
    the target when it finally stops.}
  \label{fig:results_05}
\end{figure}

Now, with \textsc{turn to light} and \textsc{forward
  locomotion} available, and with the evolution of new
photoreceptors allowed, the creature is evaluated on its
ability to navigate to a light source.  As with \textsc{turn to
  light}, fitness is averaged over multiple runs (in this case
five), again with a fixed light source at a different relative
angle each time.  The result (Figure~\ref{fig:results_05}) is
a creature whose behavioral complexity approximately matches
the previous state of the art.

\subsection*{\footnotesize \textbf{Skill 6: STRIKE}}
\label{sec:ESP_results_strike}

\begin{figure}[ht!]
\setlength{\belowcaptionskip}{-10pt}
  \includegraphics[height=1.0in]{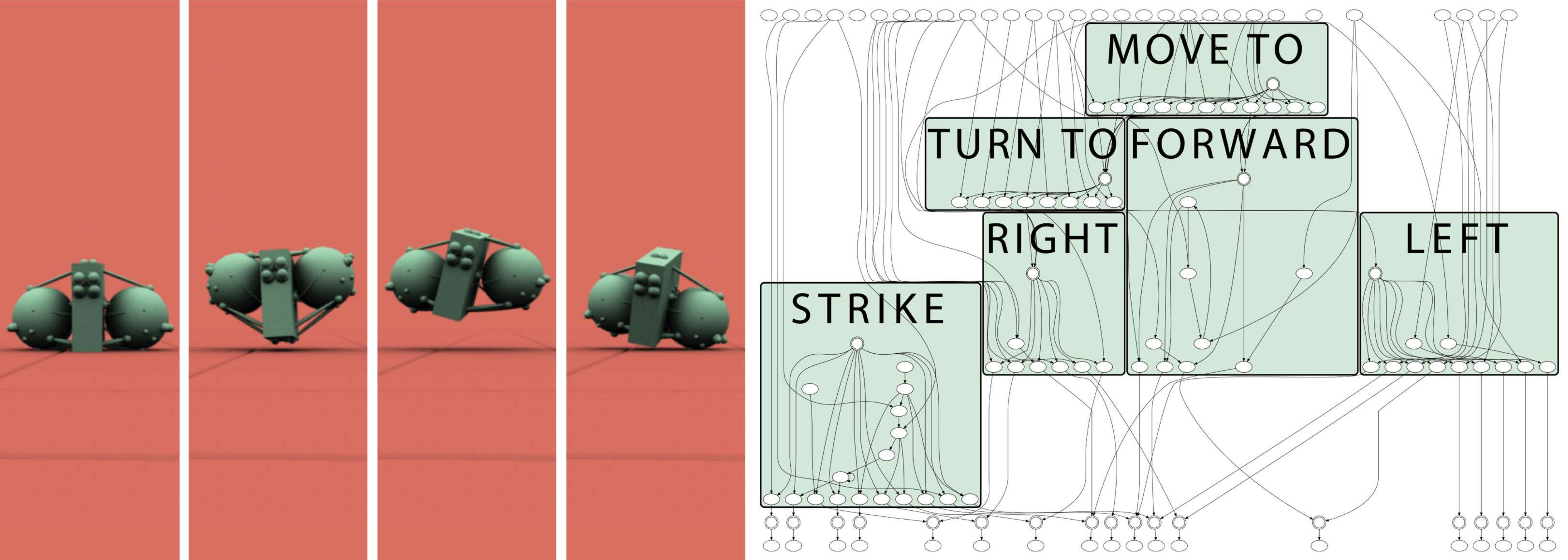}
  \caption{This creature's \textsc{strike} solution employs a
    vertical jump.}
  \label{fig:results_06}
\end{figure}

In anticipation of the upcoming \textsc{attack} task (see
Figure~\ref{fig:syllabus_graph}), the creature must first
learn to deliver a strike to the ground underneath it.  For
this creature, that goal is accomplished with a vertical jump, as seen in 
Figure~\ref{fig:results_06}.  To facilitate the evolution of
this new low-level skill, the development of new muscles is
allowed.  

\subsection*{\footnotesize \textbf{Skill 7: ATTACK (Surpassing the Previous State of the Art)}}

\begin{figure}[ht!]
\setlength{\belowcaptionskip}{-10pt}
  \includegraphics[height=1.0in]{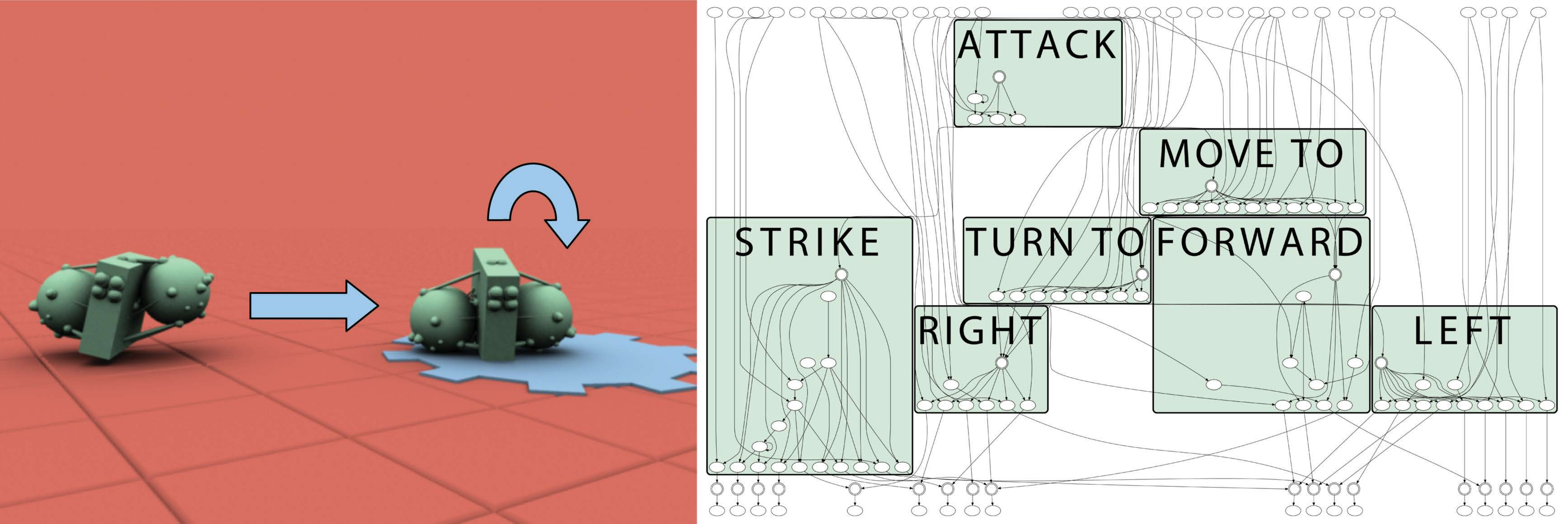}
  \caption{In the newly added \textsc{attack}, the creature
    navigates to the target, then strikes it.}
  \label{fig:results_07}
\end{figure}

Having learned \textsc{move to light} and \textsc{strike}, it
is now possible to produce an ability slightly more complex than
simply moving to a target.  By first moving to the
target, \emph{then} striking, this creature
(Figure~\ref{fig:results_07}) clearly surpasses the previous state of the art, and takes another small step
toward the behavioral complexity of compelling creature
content from the real world.  For this task, fitness is an
average across four directions of distance from the target when
the first sufficiently strong ground impact occurs (with a
penalty for producing such an impact when the scene contains
no light).

\subsection*{\footnotesize \textbf{Skill 8: TURN FROM LIGHT}}

\begin{figure}[ht!]
\setlength{\belowcaptionskip}{-10pt}
  \includegraphics[height=1.0in]{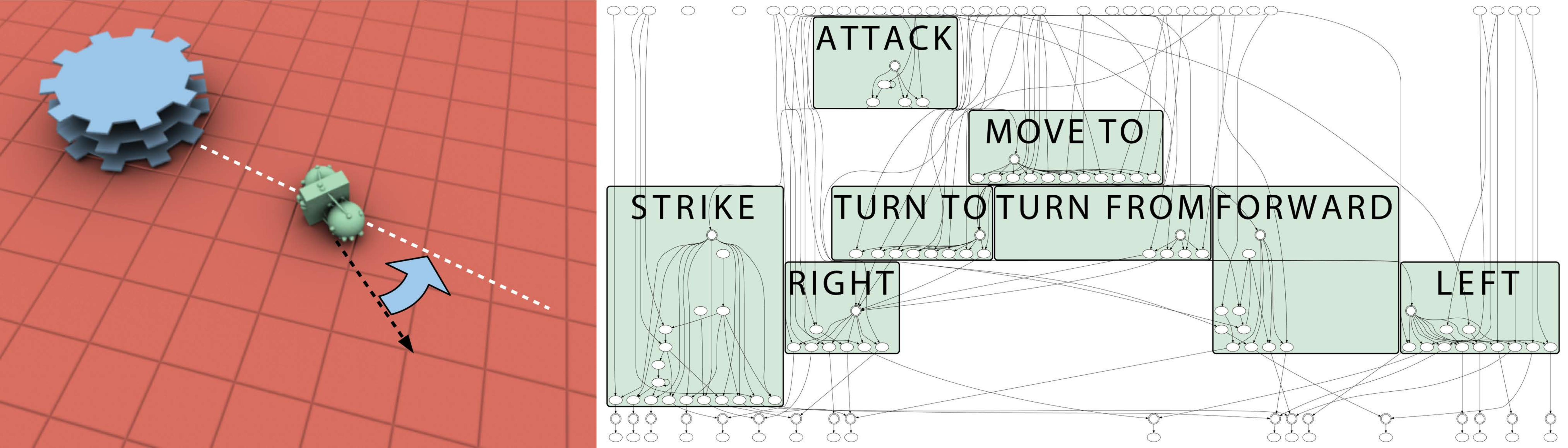}
  \caption{\textsc{turn from light} has been added, which
    keeps the locomotion direction (black dashed arrow)
    oriented away from the target.}
  \label{fig:results_08}
\end{figure}

Now, in preparation for the upcoming \textsc{retreat} skill
(see Figure~\ref{fig:syllabus_graph}), the creature must learn
to turn away from a light source
(as shown in Figure~\ref{fig:results_08}). Although obviously similar to
\textsc{turn to light}, this task also required a fitness term
to discourage an initial wrong-direction turn, so as to
achieve reasonable results for targets near the creature's
front.  Also, significantly more evaluation directions (thirteen) 
were used (particularly near the front) to similarly motivate
appropriate reactions in these cases.

\newpage
\subsection*{\footnotesize \textbf{Skill 9: RETREAT}}

\begin{figure}[ht!]
\setlength{\belowcaptionskip}{-10pt}
  \includegraphics[height=1.0in]{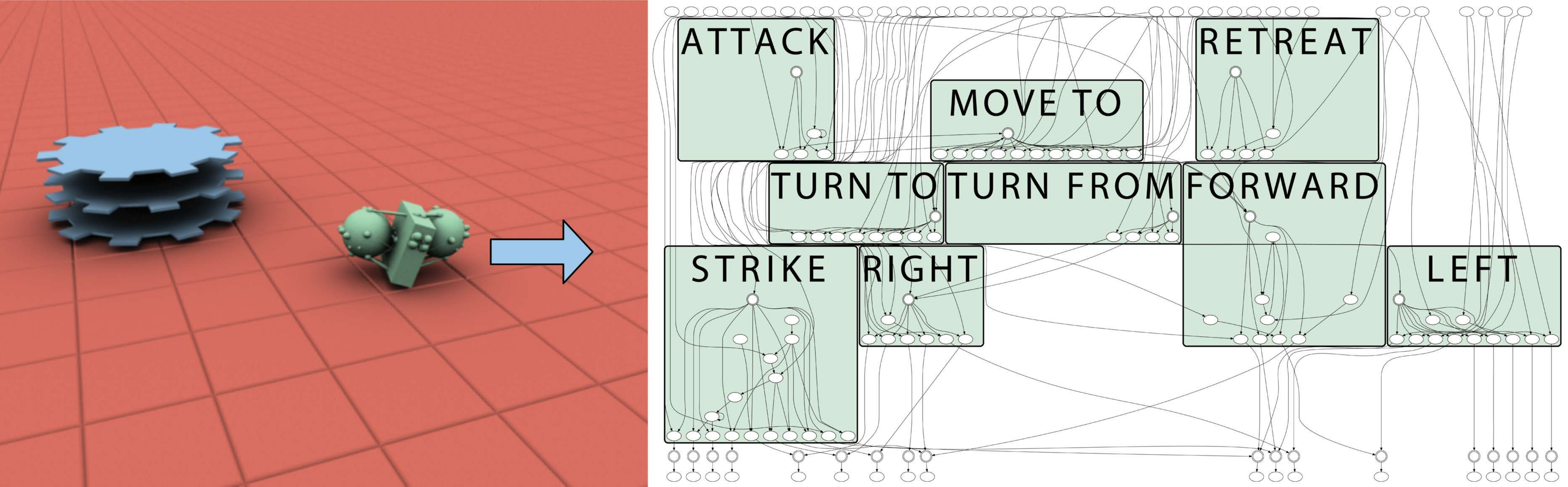}
  \caption{\textsc{retreat} added.}
  \label{fig:results_09}
\end{figure}

At this point, using \textsc{turn from light} and
\textsc{forward locomotion}, the creature learns to maximize
its average distance from a light target.  As with
\textsc{turn from light}, penalties for initial
wrong-direction moves and multiple tests with targets near the
front are combined to discourage inappropriate initial
reactions. With this skill complete
(Figure~\ref{fig:results_09}), the necessary components are in
place for the final top-level skill of the syllabus.


\subsection*{\footnotesize \textbf{Skill 10: FIGHT OR FLIGHT}}

\begin{figure}[ht!]
\setlength{\belowcaptionskip}{-10pt}
  \includegraphics[height=1.25in]{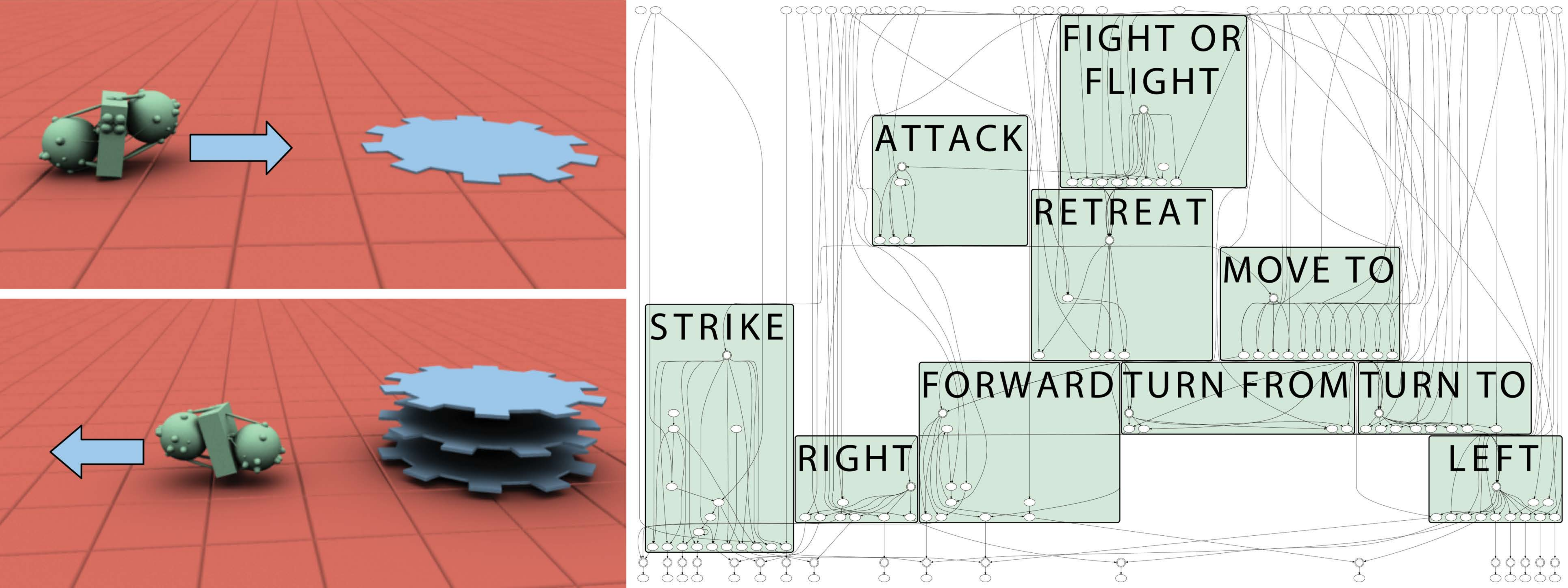}
  \caption{\textsc{fight or flight} has been added, completing
    the progression through the syllabus.}
  \label{fig:results_10}
\end{figure}

The task of this final, highest skill is to choose
between \textsc{attack} and \textsc{retreat} based on the
perceived environment.  For this evaluation, the creature is
either confronted with a vulnerable target (a single disc on the
ground), which the creature should attack, or a dangerous 
target (a spinning vertical stack of three such discs),
which will destroy the creature if touched.

The fitness score is again the result of averaging over
initial light directions, but in this case there is some
additional complexity.  At each direction, one evaluation is
made with a vulnerable target, and one with a dangerous
target.  While the proper reaction in a single case
(vulnerable vs dangerous) should
be rewarded, the real challenge is to motivate a
discrimination between the two, so that the right action can be
taken in \emph{both} cases.  To accomplish this, a small
fraction of the final score is based on the average maximum of
the two component scores (to motivate \emph{any} development,
especially initially), and a much larger fraction of the final
score is based on the average minimum of the two component
scores (to reward the ultimate goal of finding the proper
reaction in \emph{both} cases).  The weighting is chosen so
that a single perfect result for a minimum component will be
worth more than perfect scores in \emph{all} of the
maximum components.  So if $f^+$ is the average
\emph{maximum} score across all $n$ test directions, and $f^-$ is the
average \emph{minimum} score across all $n$ test directions, then the
final overall fitness is computed as
\[
\frac{f^+ + 2n \cdot f^-}{2n + 1}.
\]
Without these additional motivations, solutions emerged
which chose a single (higher-scoring) hard-coded reaction to
be used for each light position---regardless of target
type---without making the leap to the increased scores
available if discernment between the two types of target could
be developed.

Figure~\ref{fig:results_10} shows a successful
result for this task, marking the completion of the full
syllabus and the acquisition of its highest, most complex skill.
This result demonstrates that the ESP system can enable
evolved virtual creatures to achieve a level of behavioral
complexity which is a clear advance on the state of the art.


\section{General ESP}
\label{sec:extended_ESP}

The Fast ESP method achieves the goal of breaking the upper limit on behavioral complexity previously demonstrated in EVCs, but it does so at the cost of constrained morphological changes after the first skill is complete.
While that method remains useful due to its efficiency, when sufficient time or computational resources are available, it may be desirable to relax that strict morphological constraint.  The General ESP method~\cite{lessin:alife14} makes this possible, enabling full morphological adaptation to multiple tasks, while maintaining Fast ESP's ability to incrementally develop complex behaviors.

\subsection{Replacing Morphological Constraints with Retesting}
\label{sec:current_limitations}

Fast ESP enforced
strict limits on morphological changes after
the completion of the first skill:  Although changes to
muscles and photoreceptors were allowed, segments and joints
were fixed.  Due to this constraint, previously learned skills
could be expected to work reliably throughout the
syllabus-based construction.  On the other hand, this
limitation makes it difficult to develop certain abilities
later.  For example, a creature may succeed in developing
forward locomotion and the ability to turn left, but---due to
the construction of a certain joint evolved for
locomotion---be unable to learn to turn right, even after many
generations of evolution.

Luckily, when sufficient computational resources exist, this limitation 
can be removed simply by expanding and modifying the fitness
evaluations applied during learning: Instead of locking
segments and joints after the first skill is developed,
successive skills could be allowed to change these attributes,
as long as new testing shows that such changes will not
invalidate earlier abilities.

However, such an increase in testing threatens to make an already
computationally demanding problem significantly more
difficult, especially because the system is intended to be
open ended.  Assuming $n$ skills and one independent test for
each skill, full retesting of all previous skills at each step
of the syllabus would produce an $\mathcal{O}(n^2)$ growth in
the required testing, instead of Original ESP's linear growth.

\begin{figure}[t]
  \setlength{\belowcaptionskip}{-5pt}
  \centering
  \includegraphics[height=3.0in]{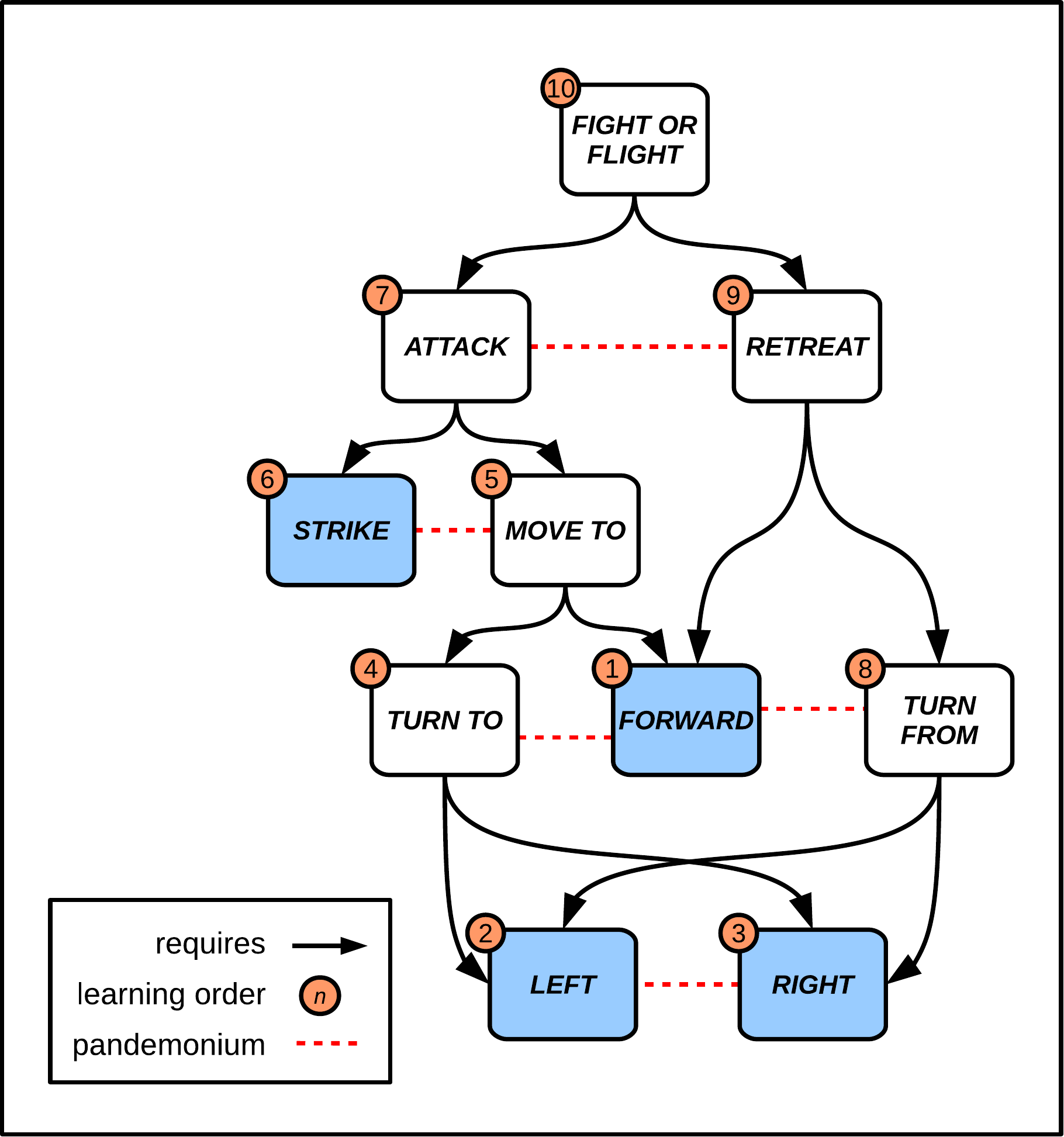}
  \caption{In this presentation of the previous section's syllabus graph, \emph{leaf nodes} (shaded) affect only the body, rather than other
nodes, and constitute the focus of the Extended ESP system.}
\label{fig:syllabus_graph_with_leaves}
\end{figure}

Fortunately, the retesting can be considerably reduced by
focusing it where it matters.  Consider again the previous syllabus
graph, as presented again in Figure~\ref{fig:syllabus_graph_with_leaves}.  The skills
that have a direct influence on the creature's body are
shaded, and will be referred to as \emph{leaf} skills. These
are: \textsc{forward locomotion}, \textsc{left turn},
\textsc{right turn}, and \textsc{strike}.  Once these skills
are successfully established, the remaining non-leaf skills
can be evolved independently (in an order that meets
dependency requirements), without the need for any retesting.
This approach stops the $\mathcal{O}(n^2)$ growth in testing
requirements significantly earlier than would otherwise be
possible--in the syllabus of Figure~\ref{fig:syllabus_graph_with_leaves},
for example, after four skills instead of 10 (assuming all
leaf skills are learned first).


\subsection{New Elements of the General ESP Algorithm}

\begin{figure}[ht!]
  \centering
  \includegraphics[width=0.45\textwidth]{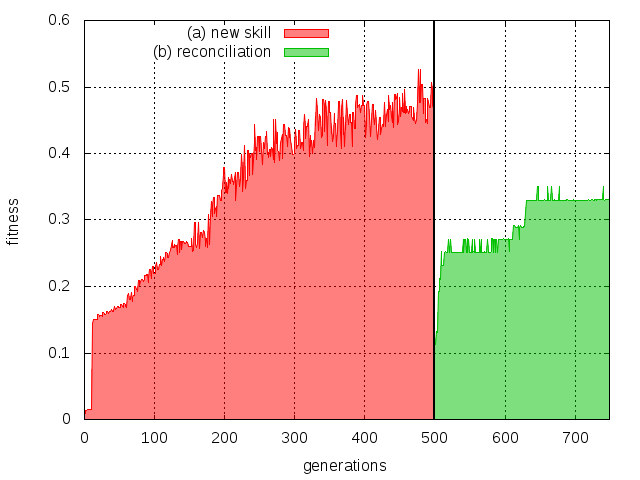}
  \caption{The fitness graph for the evolutionary run that produced the creature of Figure~\ref{fig:skitter_strike}(f).  This graph illustrates the two stages of new-skill evolution in the General ESP algorithm.  First, the new skill (in this case ~\textsc{strike}) is allowed to freely evolve both body and brain to its own ends, so long as any extant skills (in this case ~\textsc{forward locomotion}) are maintained to prescribed levels.  The fitness for the new skill during this stage is graphed as (a).  Second, morphology is preserved, while each previously developed skill is given a fixed number of generations to adapt to the new body.  The fitness for the previous skill's reconciliation to the body during this stage is graphed as (b).  These two stages work together to allow morphological adaptation for new skills, while ensuring that old skills are not lost.}
\label{fig:general_esp_fitness_graph}
\end{figure}

This section describes the elements added to Fast ESP to produce the General ESP algorithm, taking advantage of the
previous section's observation about leaf skills.  
The novel portion of General ESP is employed during the evolution of each new skill and consists of two stages.  The first
stage consists of a fixed number of generations during which
the new skill's control and body evolves, as described in
Algorithm~\ref{alg:step_1}.  During this stage, existing
encapsulated skills in the brain do not change, but if any
morphological changes reduce these skills' fitness beyond a
preset limit, the creature will be marked as unfit.  In this
way, the new skill is given free rein to adapt the body to its
needs, provided that sufficient ability in all existing skills
is retained (Figure~\ref{fig:general_esp_fitness_graph}a).

\begin{algorithm}[ht!]
  \caption{Full evolution of morphology and control for new
    skill $s'$.}
  \label{alg:step_1}
  \ForEach{generation}{
    \ForEach{individual in the population}{
      mutate morphology\;
      mutate control for new skill $s'$\;
      \ForEach{existing skill $s$}{
        evaluate fitness for $s$\;
        \If{fitness for $s$ has decreased significantly}{
          set individual fitness to 0\;
          proceed to next individual\;
        }
      }
      evaluate fitness for $s'$\;
      set individual fitness to fitness for $s'$\;
    }
    produce new population from existing one\;
  }
\end{algorithm}

\begin{algorithm}[ht!]
  \caption{Reconciling existing skills to body changes made
    for new skill $s'$.}
  \label{alg:step_2}
  \ForEach{existing skill $s$}{
    \ForEach{generation}{
      \ForEach{individual in the population}{
        mutate control for skill $s$\;
        evaluate fitness for $s$\;
        set individual fitness to fitness for $s$\;
      }
      produce new population from existing one\;
    }
  }
\end{algorithm}

The second stage (Algorithm~\ref{alg:step_2}) runs for a fixed number of generations for
each of the old skills, during which the morphology is
temporarily locked--ensuring that the abilities achieved by
the new primary skill are preserved--and each of the already
existing skills gets a chance to reconcile itself to the new
body (Figure\ref{fig:general_esp_fitness_graph}b).  Since the morphology is
fixed, these skills can develop completely independently--each
skill can adapt to the new body, without degrading any of the
other skills in the brain.

Proceeding in this manner, General ESP allows new leaf skills to seek their own adaptations
to morphology as well as control, with a reasonable
expectation that--as in the old system--existing skills will
be maintained, allowing abilities to accumulate incrementally, just
as in Fast ESP.


\section{General ESP Results}
\label{sec:extended_ESP_results}

\begin{figure}[ht!]
  \setlength{\belowcaptionskip}{-5pt}
  \centering
  \
  \begin{subfigure}{0.22\textwidth}
  \includegraphics[width=\textwidth]{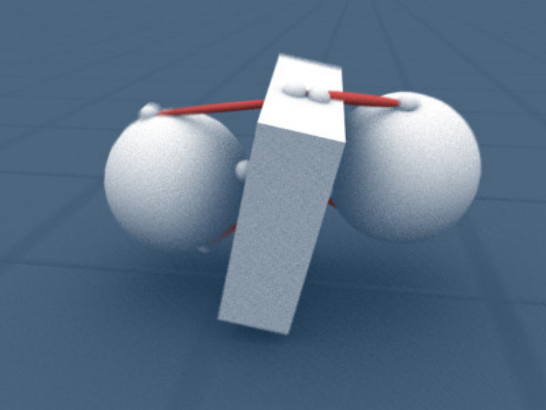}
  \caption{Initial locomoting creature.}
  \end{subfigure}
  \hspace{0.05in}
  \begin{subfigure}{0.22\textwidth}
  \includegraphics[width=\textwidth]{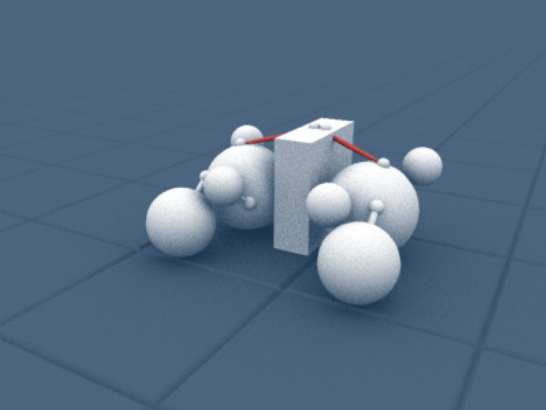}
  \caption{Heavy smashing arms.}
  \end{subfigure}

  \begin{subfigure}{0.22\textwidth}
  \includegraphics[width=\textwidth]{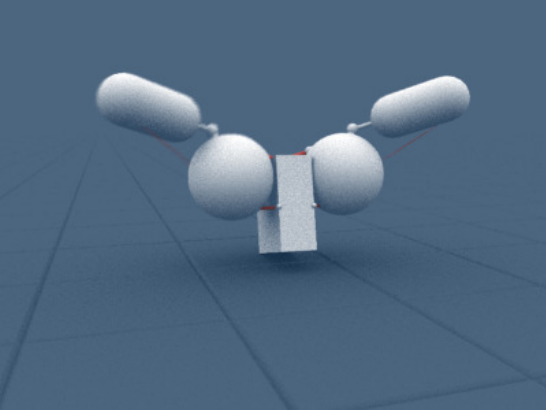}
  \caption{Smashing flail arms.}
  \end{subfigure}
  \hspace{0.05in}
  \begin{subfigure}{0.22\textwidth}
  \includegraphics[width=\textwidth]{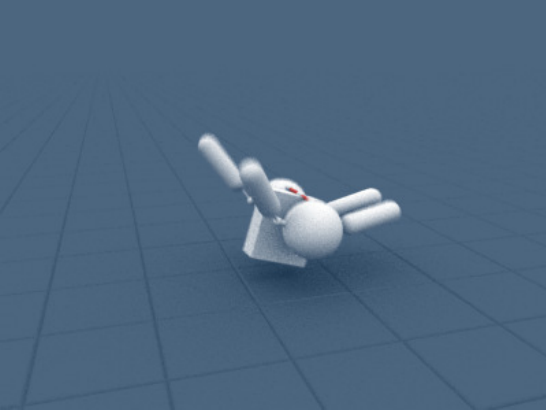}
  \caption{Jump with anti-tip limbs.}
  \end{subfigure}

  \begin{subfigure}{0.22\textwidth}
  \includegraphics[width=\textwidth]{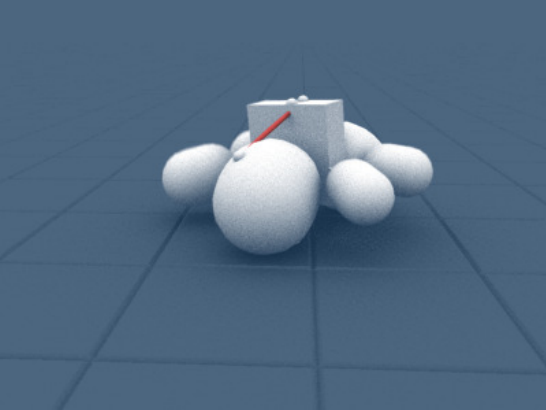}
  \caption{Smashing tail, stabilizers.}
  \end{subfigure}
  \hspace{0.05in}
  \begin{subfigure}{0.22\textwidth}
  \includegraphics[width=\textwidth]{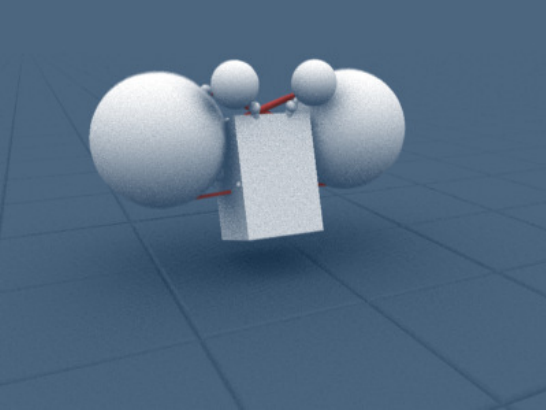}
  \caption{Jump with heavier body.}
  \end{subfigure}

  \caption{
    Adapting EVC morphology to multiple tasks.
    (a) A creature adapted for locomotion.  From this
    creature, creatures (b) through (f) were evolved using the
    General ESP method described in this paper.  Each of them
    has developed a new technique (with corresponding
    morphological changes) for accomplishing an additional
    task--in this case, delivering a strike to the
    ground--while still maintaining the ability to perform the
    initial skill (locomotion) to prescribed levels.  With Fast ESP, these secondary adaptations would
    have been impossible.
  }
  \label{fig:skitter_strike}
\end{figure}

Experiments demonstrate the advantages of the continuing
morphological evolution enabled by the General ESP
algorithm.  In the first subsection (\emph{Strike Results}),
an experiment from the Fast ESP system is reproduced in
the General ESP system, with dramatically different results.
In the second subsection (\emph{High-Reach Results}), a
learning challenge designed to highlight General ESP's
advantages is presented, and detailed benefits are described.
Note that, while General ESP maintains Fast ESP's ability
to construct complex hierarchical behaviors, that ability is
inherited largely without modification.
Therefore, the experiments in this section are used instead to
demonstrate General ESP's success in more challenging
applications that would be impossible with Fast ESP.
Video illustrating both of these result sections can be viewed
online.\footnote{http://youtu.be/fyVr7gdGEPE}
 

\subsection{Strike Results}

An important part of the Fast ESP system's primary
experimental result was to add a strike behavior to a
locomoting creature (toward the larger goal of developing a
complex fight-or-flight behavior).  In this section, that
portion of the old experiment is reproduced using General ESP, and a broad range of novel strategies and
morphological changes is observed.

\subsubsection{Strike in Fast ESP}

Figure~\ref{fig:skitter_strike}a depicts the creature evolved
for locomotion from Section~\ref{sec:ESP_results}.  Using
Fast ESP, that creature consistently solved the challenge
of producing a striking behavior by using its existing
skeletal structure to either jump up and down or smash the
ground with its limbs (Section~\ref{sec:ESP_results_strike}), without any opportunity to explore the
potential for new strategies or better adaptation that might
result from continuing full morphological development.

\subsubsection{Strike in General ESP}

When the morphology is allowed to continue to evolve, however,
new strategies become possible, and even old strategies may be
better executed with morphological changes adapted to their
specific needs.  The General ESP system develops a variety of
such solutions, as can be seen in
Figures~\ref{fig:skitter_strike}b
through~\ref{fig:skitter_strike}f.


\subsection{High-Reach Results}

The second experiment was
designed to highlight the potential benefits of General ESP
over Fast ESP.  Specifically, a selection of
three different locomoting creatures was evolved to learn the
additional skill of reaching for a high target, and the
subsequent differences of results for the Fast and
General ESP implementations were examined in detail.  General ESP led to two types of improvements: 1)
greater variety of results, and 2) better
fitness.

\subsubsection{Greater Variety}

\begin{figure}[ht]
  \centering
  \
  \begin{subfigure}{0.22\textwidth}
  \includegraphics[width=\textwidth]{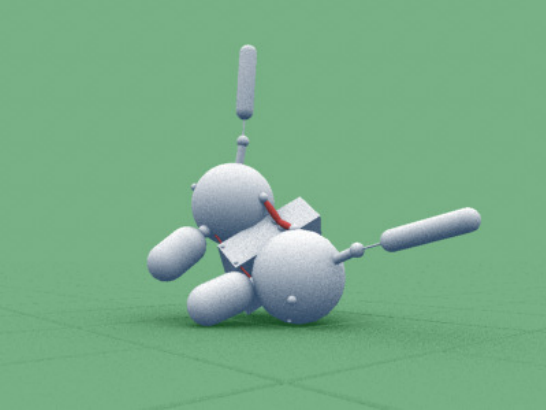}
  \caption{Tipping, long new limbs.}
  \end{subfigure}
  \hspace{0.05in}
  \begin{subfigure}{0.22\textwidth}
  \includegraphics[width=\textwidth]{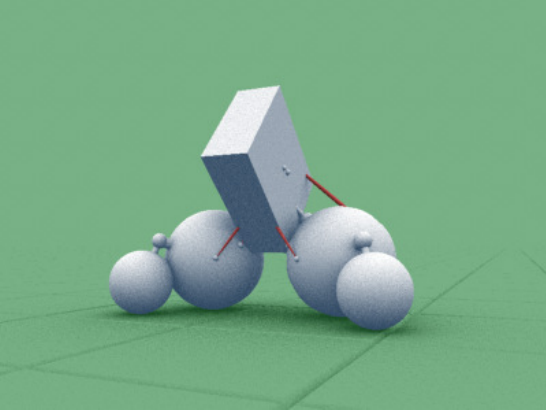}
  \caption{Push-up, extended limbs.}
  \end{subfigure}

  \begin{subfigure}{0.22\textwidth}
  \includegraphics[width=\textwidth]{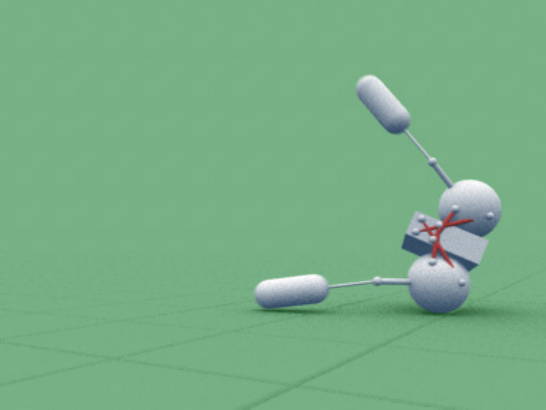}
  \caption{Telescoping limbs.}
  \end{subfigure}
  \hspace{0.05in}
  \begin{subfigure}{0.22\textwidth}
  \includegraphics[width=\textwidth]{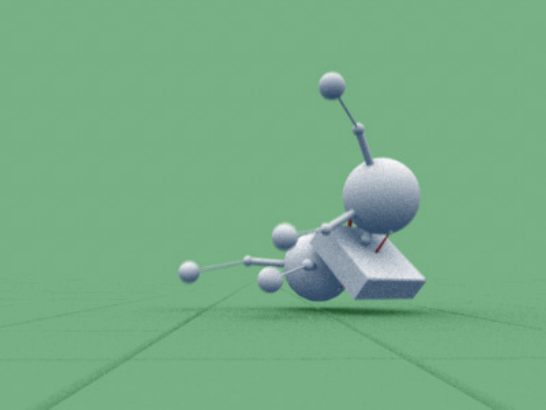}
  \caption{Telescoping, anti-tip limbs.}
  \end{subfigure}

  \begin{subfigure}{0.22\textwidth}
  \includegraphics[width=\textwidth]{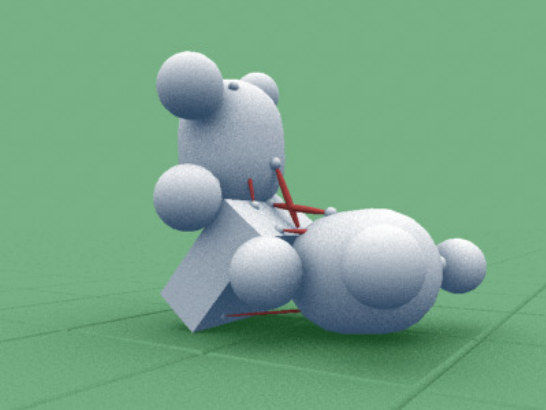}
  \caption{Tip with enlarged limbs.}
  \end{subfigure}
  \hspace{0.05in}
  \begin{subfigure}{0.22\textwidth}
  \includegraphics[width=\textwidth]{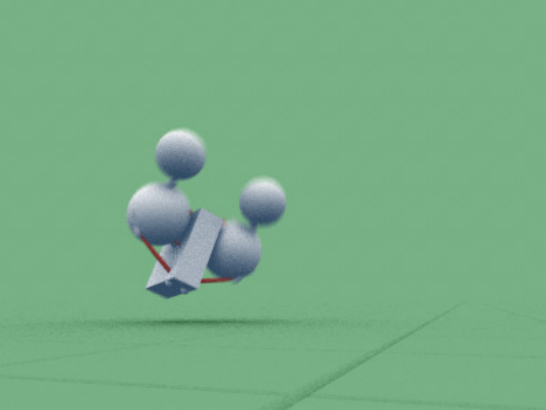}
  \caption{Jump, swing extensions up.}
  \end{subfigure}

  \caption{
    Greater variety through continued morphology evolution.
    The locomoting creature of
    Figure~\ref{fig:skitter_strike}a was further evolved using
    the General ESP system to adapt to a high-reach task.  The
    results demonstrate the potential of continued morphology
    evolution to produce a great degree of useful variety.
  }
  \label{fig:skitter_high}
\end{figure}

The locomoting creature of Figure~\ref{fig:skitter_strike}a
was evolved toward the new high-reach goal, using both the
Fast and the General ESP implementations.

With fast ESP, only two strategies were observed,
within which the results were extremely uniform.  Using
morphology unchanged from the original locomotion result, all
such creatures developed to either jump as high as possible,
or reach one limb up by tipping over onto the other limb.  In
both cases, the results were limited by the inability of
skeletal morphology to adapt to this new task.

With General ESP, in contrast, a wide
variety of results was observed, in which a number of novel
strategies were used, often to great effect.  These solutions
are illustrated in Figure~\ref{fig:skitter_high} (a) through
(f).

\subsubsection{Better Fitness}

\begin{figure}[t]
  \setlength{\belowcaptionskip}{-5pt}
  \centering
  \
  \begin{subfigure}{0.22\textwidth}
  \includegraphics[width=\textwidth]{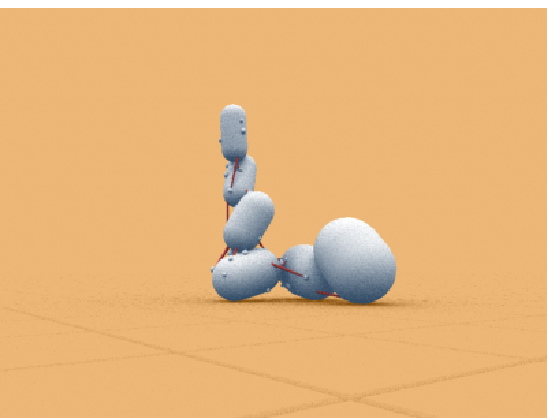}
  \caption{Original ESP result.}
  \end{subfigure}
  \hspace{0.05in}
  \begin{subfigure}{0.22\textwidth}
  \includegraphics[width=\textwidth]{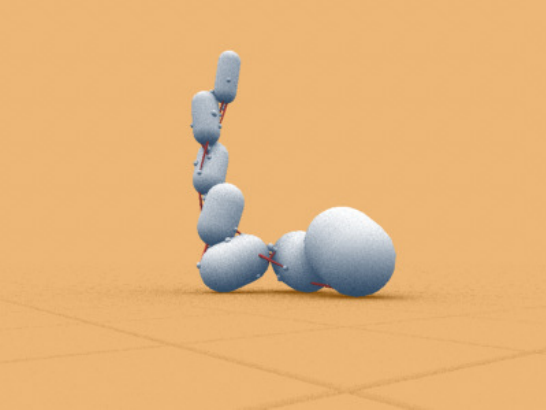}
  \caption{Result in new ESP.}
  \end{subfigure}

  \caption{
    Improved fitness via continued morphology evolution.
    These results demonstrate how the General ESP system (b)
    can produce better fitness values (i.e., a higher reach)
    than the Fast ESP system (a) by allowing the addition
    of new body segments.
  }
  \label{fig:snake_high}
\end{figure}

Another successful solution to the locomotion task produced
by Fast ESP is shown in Figure~\ref{fig:snake_high}a.
This snake-like creature achieved a high reach by extending
one end of its long morphology (best fitness in 10 runs: 0.174), while the rest of
the body maintained balance.

General ESP improved upon this creature by changing its
morphology for the secondary task, while its strategy remained
unchanged (Figure~\ref{fig:snake_high}b).  With General ESP, the creature was able to develop an
additional body segment that enabled the higher reach (fitness 0.267), while
allowing it still to perform locomotion to acceptable
standards.

\subsubsection{Greater Variety \emph{and} Better Fitness}

\begin{figure}[ht]
  \centering

  \begin{subfigure}{0.22\textwidth}
  \includegraphics[width=\textwidth]{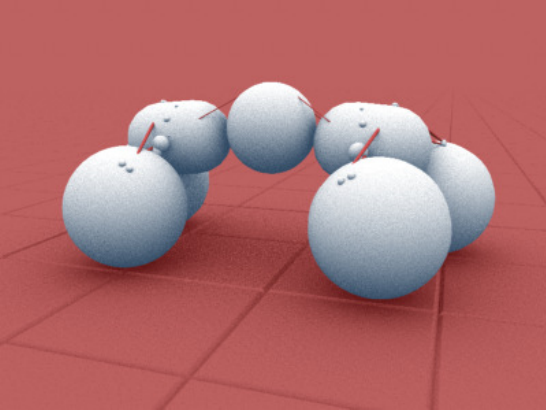}
  \caption{Initial locomoting creature.}
  \end{subfigure}
  \hspace{0.05in}
  \begin{subfigure}{0.22\textwidth}
  \includegraphics[width=\textwidth]{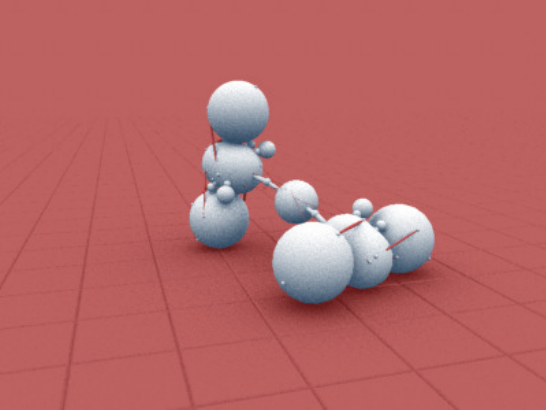}
  \caption{Subtle body changes.}
  \end{subfigure}

  \begin{subfigure}{0.22\textwidth}
  \includegraphics[width=\textwidth]{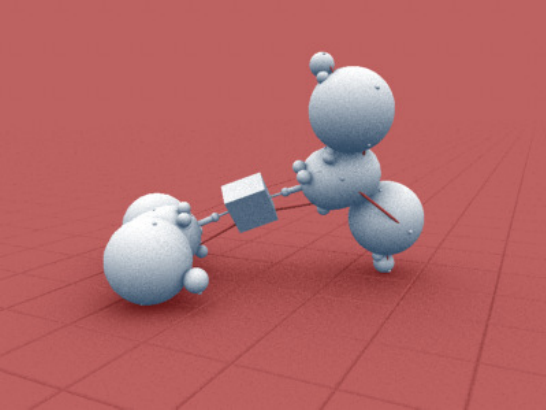}
  \caption{More obvious body changes.}
  \end{subfigure}
  \hspace{0.05in}
  \begin{subfigure}{0.22\textwidth}
  \includegraphics[width=\textwidth]{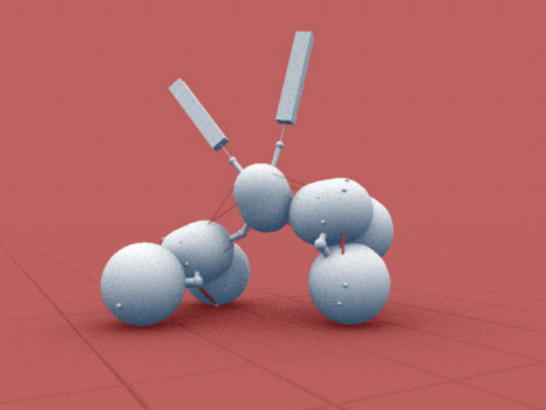}
  \caption{Dramatic changes in morphology.}
  \end{subfigure}

  \caption{
    Greater variety and improved fitness.  The initial
    locomoting quadruped (a) is evolved for high reach in the
    General ESP system (b)-(d).  Through a variety of
    strategies, each of the General ESP creatures scores
    better on this new task than any creature from the
    Fast ESP system.
  }
  \label{fig:atat_high}
\end{figure}

The relatively complex quadruped seen in
Figure~\ref{fig:atat_high}a was a third solution
developed by the underlying EVC system for the locomotion
task.  In continued evolution of the high-reach task in the
Fast ESP system, this creature's results were again
extremely uniform in approach and fitness.  They all reached
up with a single limb, and all with approximately equal
success (best fitness in 10 runs: 0.164).  In the General ESP system, the ability to continue to
adapt morphology to this new task led to a diverse set of
useful results, with all presented being more fit than any
produced with Fast ESP.

Figure~\ref{fig:atat_high}b depicts a creature
that pursues the same strategy as the creature in
Figure~\ref{fig:atat_high}a, yet does so more effectively (fitness 0.209)
due to subtle morphological adaptations.  In
Figure~\ref{fig:atat_high}c, more obvious morphological
adaptations have been added to further exceed
the uniform performance limit experienced by this creature in
Fast ESP, while still employing the same basic
technique (fitness 0.294).  In Figure~\ref{fig:atat_high}d, even
more dramatic changes to morphology provide a new way of
solving the high reach: This creature (fitness 0.314) employs a new pair of tall,
dedicated limbs to even further exceed the performance of Fast ESP.


\section{Discussion and Future Work}

In this section, outstanding issues related to ESP are discussed, and potential avenues for future development are presented.

\subsection{ESP's Requirement for Human Input}

While it is true that some human input is required by the ESP system,
it is important to note that
the human input utilized by this method in the form of the
syllabus is at a usefully abstract level---on a par with the
kind of input employed by human learners.  This syllabus,
along with the opportunity for human selection among high
scorers at the end of each subskill stage, offers great
potential value as a mechanism for exerting relatively
high-level creative control over creature development.
In addition, specifying even a single fitness function in a traditional EVC system  arguably places even greater demands on the human experimenter than the creation of such a syllabus.

\subsection{Benefits of Evolving Creature Content}

Numerous benefits accrue from the fact that this
system's results are evolved and that this evolution takes
place in physical simulation.  Thanks to evolution, the
creatures this system produces are unceasingly novel,
developing new solutions for morphology, muscle and eye
placement, and mechanism and style of movement each time the
process is restarted.  And the fact that these solutions
are evolved to operate in a physically simulated environment
adds a particular level of realism, demonstrating results that
are convincingly physically plausible, and even include some
of the subtle imperfections of action that bring so much
character to creatures in the real world.  Note, also, that
creating controllers for bodies like these by hand would
be impractical, but that this difficulty is in this case
handled entirely by the evolutionary algorithm.

\subsection{ESP is Open-Ended}

One of the most important aspects of the ESP system is
that it is designed to be open ended.  While a significant
increase in behavioral complexity has been demonstrated, there
are no obvious barriers to continued reapplication of this
technique to achieve results of still greater complexity in
the future.
Regardless of the work it
took to achieve the top-level fight-or-flight behavior
described above, once complete and encapsulated, that entire
skill can be easily utilized as a unit by future evolution.
For example, it might next be useful to add a tip-crisis behavior:
Any time the creature finds itself tipped over, it would work to
right itself before continuing.  This tip-recovery action
could be learned by a creature which has completed the example
fight-or-flight syllabus above, then a new top-level ability
could be evolved that simply chooses between tip-recovery and
normal fight-or-flight behaviors based on whether or not the
creature is upright.  The ESP system is designed to make this step
(and those beyond it) equally straightforward to evolve
solutions to.

\subsection{Beyond General ESP}

Although the General ESP algorithm has removed Fast ESP's explicit limitations on body changes after the first
skill, development of morphology throughout the acquisition of
complex skills is still not fully general and completely
unlimited.
First, the retesting requirements would make morphological
development impractical if continued through too many steps of
leaf skill addition.  To mitigate this issue in the future, it
may be possible to do the retesting periodically rather than
universally, and run the tests in parallel.
Also, the more leaf skills there are, the more likely it is
that the morphological change required by one skill is harmful
to the others.  This limitation is more difficult to
overcome, and indeed it reflects the conflicting demands that
any creature faces when dealing with complex environments.  

\subsection{Decomposition of Perception}

Just as the ESP system
decomposes complex actions into simpler
ones for piece-by-piece learning, an analogous process might
decompose perception for a similar benefit.  As part of a more
complex syllabus, a human expert could develop a sequence of
sensing tasks leading to useful perceptual abilities that
might be difficult or impossible to achieve otherwise.  This,
in turn, could make possible greater behavioral abilities
overall.

\subsection{Combat}

While Miconi has already produced one limited form of
combat for EVCs~\cite{miconi2008silicon}, there is a
great deal more that can be done in this area.  The ESP
method, in combination with the future-work topics 
described above (and the ability to vary body-part materials, the
importance of which was recognized by Miconi), could
potentially produce a far richer and more compelling form of
combat for evolved virtual creatures than what has been seen to
date.

\subsection{Fauna on Demand}

Finally, a more refined and automated
version of the ESP system could make it
possible to populate virtual worlds with continually novel
creature content (especially with the help of techniques such
as those seen in~\cite{lehman2011evolving}).  As virtual
boundaries are pushed back, human users could (subject to
limitations of computing power) continually encounter
never-before-seen creatures, all developed from a single
high-level human-designed syllabus.

\bigskip 

Future work examples such as these illustrate that, beyond the demonstrated advances due to ESP, significant avenues of new research are made possible by this technique's introduction.


\section{Conclusion}

The ESP system described in this paper allows evolved
virtual creatures to achieve a level of 
behavioral complexity (as defined in the Introduction) which is approximately double the state
of the art.  
In contrast to related techniques for fixed morphologies, this advance applies when morphology is evolved as well as control, demonstrating the first clear increase for that application in the past two decades. 

Two versions of the ESP system were presented, Fast and General, distinguished by
the relative importances of computation time and extended morphological adaptation.
In exchange for some limitations on continuing morphological changes, Fast ESP makes it possible to increase behavioral complexity with only a linear increase in computational time.  When computational resources permit it, General ESP in contrast makes it possible to adapt morphology fully beyond the initial skill.  It results in a greater variety of solutions and solutions
with higher fitness, while still permitting the same open-ended development of complex behaviors as Fast ESP.

These advances demonstrate that the potential for behavioral
complexity in evolved virtual creatures has not yet been
exhausted, and in fact suggests that it may continue to
increase so as to one day match the behavioral complexity of
creatures from the real world---with all of the promise for
content creation that such complexity might bring.






\section*{Acknowledgments}

This research was supported by NSF grants DBI-0939454
and IIS-0915038, and by equipment donations from Intel’s
Visual Computing Program.

\ifCLASSOPTIONcaptionsoff
  \newpage
\fi



\bibliographystyle{abbrv}
\bibliography{IEEEabrv,dissertation}

\end{document}